\documentclass[10pt,twocolumn,letterpaper]{article}

\usepackage{cvpr}
\usepackage{times}
\usepackage{epsfig}
\usepackage{graphicx}
\usepackage{amsmath}
\usepackage{subfig}
\usepackage{amssymb}
\usepackage{tabularx}
\usepackage{booktabs}
\usepackage{multirow}
\usepackage{multicol}
\usepackage{csquotes}
\usepackage{rotating}
\usepackage{floatrow}
\usepackage{array, makecell}

\usepackage[pagebackref=true,breaklinks=true,letterpaper=true,colorlinks,bookmarks=false]{hyperref}

\hypersetup{
  colorlinks, linkcolor=red
}
\makeatletter
\g@addto@macro{\UrlBreaks}{\UrlOrds}
\makeatother

\usepackage{capt-of,etoolbox}

\newcommand{\norm}[1]{\left\lVert#1\right\rVert}
\DeclareMathOperator{\Warp}{Warp}
\newcommand{\bx}{\mathbf{x}}
\newcommand{\bxp}{\mathbf{x}_p}
\newcommand{\bxc}{\mathbf{x}_c}
\newcommand{\bs}{\mathbf{s}}
\newcommand{\IR}{\mathbb{R}}
\newcommand{\IE}{\mathbb{E}}
\newcommand{\EL}{\mathcal{L}}

\DeclareRobustCommand{\etal}{\textit{et al.}\@\xspace}
\renewcommand{\mkbegdispquote}[2]{\itshape}

\cvprfinalcopy % *** Uncomment this line for the final submission

\def\cvprPaperID{6807} % *** Enter the CVPR Paper ID here

\ifcvprfinal\fi
\begin{document}
%%%%%%%%% TITLE
\title{WarpGAN: Automatic Caricature Generation}

\author{Yichun Shi\footnotemark[1]\quad\quad Debayan Deb\footnotemark[1]\quad\quad Anil K. Jain\\
Michigan State University, East Lansing MI 48824\\
{\tt\small \{shiyichu, debdebay\}@msu.edu, jain@cse.msu.edu}
}

\twocolumn[{%
\renewcommand\twocolumn[1][]{#1}%
\maketitle
\begin{center}
\vspace{-0.8em}
    \centering
    \captionsetup{font=small}
    \begin{minipage}{0.18\linewidth}
    \includegraphics[width=\linewidth]{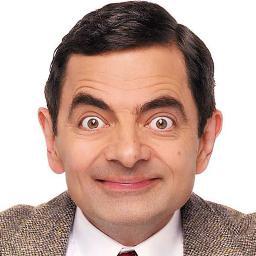}\\[0.1em]
    \includegraphics[width=\linewidth]{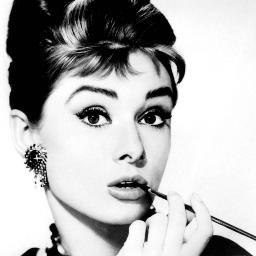}\\
    \centering {\small(a) Photo}
    \end{minipage}\;
    \begin{minipage}{0.18\linewidth}
    \includegraphics[width=\linewidth]{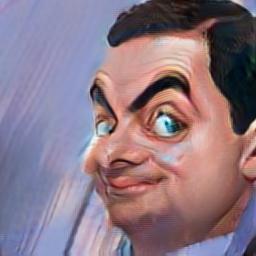}\\[0.1em]
    \includegraphics[width=\linewidth]{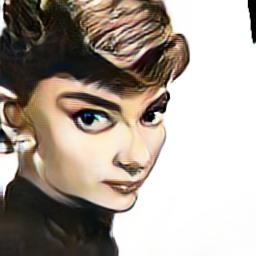}\\
    \centering {\small(b) WarpGAN}
    \end{minipage}\;
    \begin{minipage}{0.18\linewidth}
    \includegraphics[width=\linewidth]{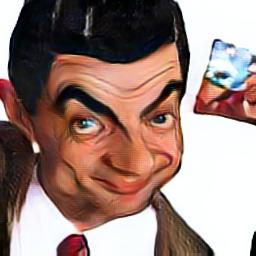}\\[0.1em]
    \includegraphics[width=\linewidth]{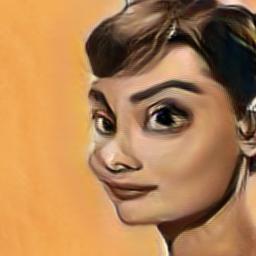}\\
    \centering {\small(c) WarpGAN}
    \end{minipage}\;
    \begin{minipage}{0.18\linewidth}
    \includegraphics[width=\linewidth]{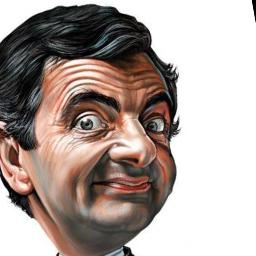}\\[0.1em]
    \includegraphics[width=\linewidth]{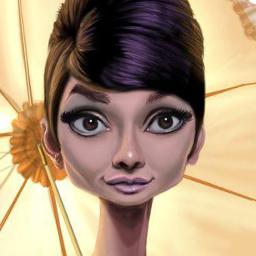}\\
    \centering {\small(d) Artist}
    \end{minipage}\;
    \begin{minipage}{0.18\linewidth}
    \includegraphics[width=\linewidth]{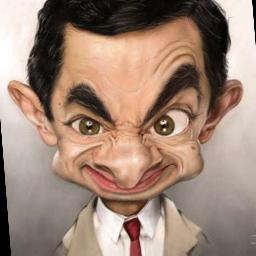}\\[0.1em]
    \includegraphics[width=\linewidth]{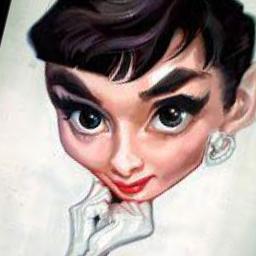}\\
    \centering {\small(e) Artist}
    \end{minipage}
    \vspace{-0.8em}\captionof{figure}{Example photos and caricatures of two subjects in our dataset. Column (a) shows each identity's real face photo, while two generated caricatures of the same subjects by WarpGAN are shown in column (b) and (c). Caricatures drawn by artists are shown in the column (d) and (e).}
    \label{fig:frontpage}
\end{center}%
}]

\thispagestyle{empty}

% \twocolumn[{%
% \renewcommand\twocolumn[1][]{#1}%
% \maketitle
% \thispagestyle{empty}
% \begin{center}
%     \centering
%     \subfigure{
%     \includegraphics[width=0.15\linewidth]{fig/fronpage/10.png}
%     \includegraphics[width=0.15\linewidth]{fig/fronpage/20.png}}
%     \subfigure{
%     \includegraphics[width=0.15\linewidth]{fig/fronpage/11.png}
%     \includegraphics[width=0.15\linewidth]{fig/fronpage/21.png}}
%     \subfigure{
%     \includegraphics[width=0.15\linewidth]{fig/fronpage/12.png}
%     \includegraphics[width=0.15\linewidth]{fig/fronpage/22.png}}
%     \subfigure{
%     \includegraphics[width=0.15\linewidth]{fig/fronpage/13.jpg}
%     \includegraphics[width=0.15\linewidth]{fig/fronpage/23.jpg}}
%     \subfigure{
%     \includegraphics[width=0.15\linewidth]{fig/fronpage/14.jpg}
%     \includegraphics[width=0.15\linewidth]{fig/fronpage/24.jpg}}
%     \captionof{figure}{Example photos and caricatures of two subjects in our dataset. Each identity's real face photo is shown in the first column, while two generated caricatures from WarpGAN are shown in the subsequent two columns. Caricatures drawn by artists of the same subject are shown in the last two columns.}
%     \label{fig:frontpage}
% \end{center}%
% }]

%%%%%%%%% ABSTRACT
\begin{abstract}
\vspace{-0.5em}
   We propose, WarpGAN, a fully automatic network that can generate caricatures given an input face photo. Besides transferring rich texture styles, WarpGAN learns to automatically predict a set of control points that can warp the photo into a caricature, while preserving identity. We introduce an identity-preserving adversarial loss that aids the discriminator to distinguish between different subjects. Moreover, WarpGAN allows customization of the generated caricatures by controlling the exaggeration extent and the visual styles. Experimental results on a public domain dataset, WebCaricature, show that WarpGAN is capable of generating caricatures that not only preserve the identities but also outputs a diverse set of caricatures for each input photo. Five caricature experts suggest that caricatures generated by WarpGAN are visually similar to hand-drawn ones and only prominent facial features are exaggerated.
\end{abstract}

%%%%%%%%% BODY TEXT
\vspace{-1.0em}
\section{Introduction}

A \emph{caricature} is defined as ``\textit{a picture, description, or imitation of a person or a thing in which certain striking characteristics are exaggerated in order to create a comic or grotesque effect}''~\cite{caricature_definition}. Paradoxically, caricatures are images with facial features that represent the face more than the face itself. Compared to cartoons, which are 2D visual art that try to re-render an object or even a scene in a usually simplified artistic style, caricatures are portraits that have exaggerated features of a certain persons or things. Some example caricatures of two individuals are shown in Figure~\ref{fig:frontpage}. The fascinating quality of caricatures is that even with large amounts of distortion, the identity of person in the caricature can still be easily recognized by humans. In fact, studies have found that we can recognize caricatures even more accurately than the original face  images~\cite{rhodes1987identification}. 
% \begin{displayquote}
% However regular we may imagine a face to be, however harmonious its lines and supple its movements, their adjustment is never altogether perfect: there will always be discoverable the signs of some impending bias, the vague suggestion of a possible grimace, in short some favourite distortion towards which nature seems to be particularly inclined. The art of the caricaturist consists in detecting this, at times, imperceptible tendency, and in rendering it visible to all eyes by magnifying it. --Henri Bergson, 1900~\cite{henri_bergson}
% \end{displayquote}

{
  \renewcommand{\thefootnote}%
    {\fnsymbol{footnote}}
  \footnotetext[1]{\;indicates equal contribution}
}

Caricature artists capture the most important facial features, including the face and eye shapes, hair styles, etc. Once an artist sketches a rough draft of the face, they will start to exaggerate person-specific facial features towards a larger deviation from an average face. Nowadays, artists can create realistic caricatures through computer softwares through: (1) warping the face photo to exaggerate the shape and (2) re-rendering the texture style~\cite{photoshop}. By mimicking this process, researchers have been working on automatic caricature generation~\cite{liang2002example,lewiner2011interactive}. A majority of the studies focus on designing a good structural representation to warp the image and change the face shape. However, neither the identity information nor the texture differences between a caricature and a face photo are taken into consideration. In contrast, numerous works have made progress with deep neural networks to transfer image styles~\cite{unit,MUNIT}. Still these approaches merely focus on translating the texture style forgoing any changes in the facial features.

\begin{table*}[!t]
    \centering
    \captionsetup{font=small}
    \caption{Comparison of various studies on caricature generation. Majority of the published studies focus on either deforming the faces or transferring caricature styles, unlike the proposed WarpGAN which focuses on both. On the other hand, WarpGAN deforms the face in the image space thereby, truly capturing the transformations from a real face photo to a caricature. Moreover, WarpGAN does not require facial landmarks for generating caricatures.}
    \label{tab:comparison_works}
    \resizebox{\textwidth}{!}{
        \begin{tabular}{|c|ccc|c|}
        \noalign{\hrule height 1.5pt}
        \rule{0pt}{2.5ex}\textbf{Approach} & \multicolumn{3}{c|}{\textbf{Methodology}} & \textbf{Examples}\\ \cline{2-4}
             \rule{0pt}{2.5ex} & \textbf{Study} & \textbf{Exaggeration Space} & \textbf{Warping}  &\\
            \noalign{\hrule height 1.2pt}
            \multirow{4}{*}{Shape Deformation} & & & & \multirow{4}{*}{
            \begin{minipage}{0.5\linewidth}\vspace{0.5em}
                \begin{minipage}{0.25\linewidth}
                    \centering  \includegraphics[width=1.7cm]{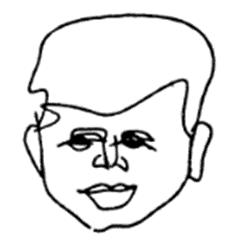}\\
                    \centering \cite{brennan1985caricature}
                \end{minipage}\hfill
                \begin{minipage}{0.25\linewidth}
                    \includegraphics[width=1.7cm]{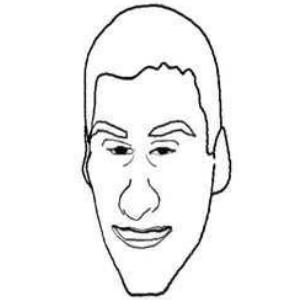}\\
                    \centering \cite{liang2002example}
                \end{minipage}\hfill
                \begin{minipage}{0.5\linewidth}
                    \includegraphics[width=1.7cm]{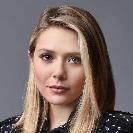}
                    \includegraphics[width=1.7cm]{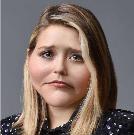}\\
                    \centering \cite{han2018caricatureshop}
                \end{minipage}
            \end{minipage}}\\
            & Brennan~\etal~\cite{brennan1985caricature} & Drawing Line & User-interactive  &\\[0.8em] 
            & Liang~\etal~\cite{liang2002example} & 2D Landmarks & User-interactive  &\\[0.8em] 
             & CaricatureShop~\cite{han2018caricatureshop} & 3D Mesh & Automatic &\\[0.6em]
             \hline
             \multirow{4}{*}{Texture Transfer} & & & & \multirow{4}{*}{
            \begin{minipage}{0.5\linewidth}\vspace{0.5em}
                \begin{minipage}{0.5\linewidth}
                    \includegraphics[width=1.7cm]{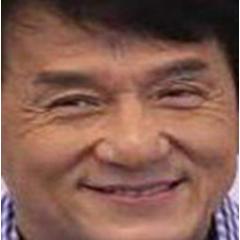}
                    \includegraphics[width=1.7cm]{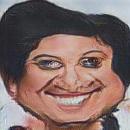}\\
                    \centering \cite{zheng2017photo}
                \end{minipage}\hfill
                \begin{minipage}{0.5\linewidth}
                    \includegraphics[width=1.7cm]{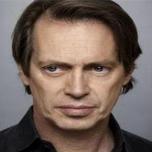}
                    \includegraphics[width=1.7cm]{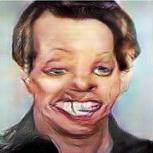}\\
                    \centering \cite{carigan}
                \end{minipage}
            \end{minipage}}\\[0.6em]
            & Zheng~\etal~\cite{zheng2017photo} & Image to Image & None &\\[0.8em]
            & CariGAN~\cite{carigan} & Image + Landmark Mask & None &\\[0.8em]
            & & & &\\
             \hline
       \multirow{4}{*}{Texture + Shape} & & & & \multirow{4}{*}{
            \begin{minipage}{0.5\linewidth}\vspace{0.5em}
                \begin{minipage}{0.5\linewidth}
                    \includegraphics[width=1.7cm]{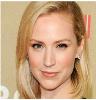}
                    \includegraphics[width=1.7cm]{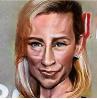}\\
                    \centering \cite{carigans}
                \end{minipage}\hfill
                \begin{minipage}{0.5\linewidth}
                    \includegraphics[width=1.7cm]{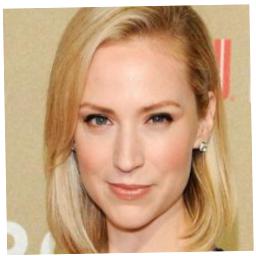}
                    \includegraphics[width=1.7cm]{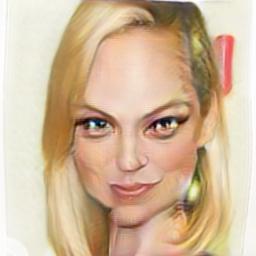}\\
                    \centering Ours
                \end{minipage}
            \end{minipage}}\\[0.7em]
            & CariGANs~\cite{carigans} & PCA Landmarks & Automatic &\\[0.8em]
            & WarpGAN & Image to Image & Automatic&\\[0.8em]
            & & & &\\
     \noalign{\hrule height 1.5pt}
    \end{tabular}
    }
\end{table*}

In this work, we aim to build a completely automated system that can create new caricatures from photos by utilizing Convolutional Neural Networks (CNNs) and Generative Adversarial Networks (GANs). Different from previous works on caricature generation and style transfer, we emphasize the following challenges in our paper:
\begin{itemize}
    \itemsep0.2em
    \item The caricature generation involves both texture changes and shape deformation.
    \item The faces need to be exaggerated in a manner such that they can still be recognized.
    \item Caricature samples exist in various visual and artistic styles (see Figure~\ref{fig:frontpage}). 
\end{itemize}

In order to tackle these challenges, we propose a new type of style transfer network, named WarpGAN, which decouples the shape deformation and texture rendering into two tasks. Akin to a human operating an image processing software, the generator in our system automatically predicts a set of control points that warp the input face photo into the closest resemblance to a caricature and also transfers the texture style through non-linear filtering. The discriminator is trained via an identity-preserving adversarial loss to distinguish between different identities and styles, and encourages the generator to synthesize diverse caricatures while automatically exaggerating facial features specific to the identity. Experimental results show that compared to state-of-the-art generation methods, WarpGAN allows for texture update along with face deformation in the image space, while preserving the identity. Compared to other style transfer GANs~\cite{cycle_gan_style_transfer,MUNIT}, our method not only permits a transfer in texture style, but also deformation in shape. The contributions of the paper can be summarized as follows:
\begin{itemize}
    \itemsep0.2em
    \item A domain transfer network that decouples the texture style and geometric shape by automatically estimating a set of sparse control points to warp the images.
    \item A joint learning of texture style transfer and image warping for domain transfer with adversarial loss. 
    \item A quantitative evaluation through face recognition performance shows that the proposed method retains identity information after transferring texture style and warping. In addition, we conducted two perceptual studies where five caricature experts suggest that WarpGAN generates caricatures that are (1) visually appealing, (2) realistic; where only the appropriate facial features are exaggerated, and (3) our method outperforms the state-of-the-art.
    \item An open-source\footnote{\url{https://github.com/seasonSH/WarpGAN}} automatic caricature generator where users can customize both the texture style and exaggeration degree. 
\end{itemize}\vspace{-0.2em}

\begin{figure}[!t]
\captionsetup{font=small}
\subfloat[Global Parameters~\cite{jaderberg2015spatial}~\cite{dong2018softgated}~\cite{lin2018stgan}]{\includegraphics[width=0.49\linewidth]{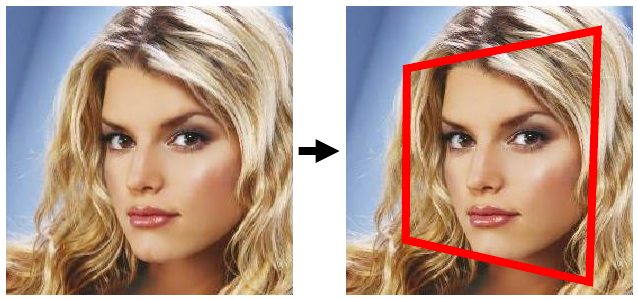}}\hfill
\subfloat[Dense Deformation Field~\cite{ganin2016deepwarp}]{\includegraphics[width=0.49\linewidth]{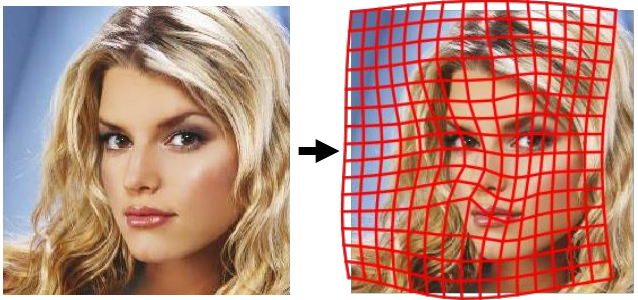}}\\
\subfloat[Landmark-based~\cite{cole2017synthesizing}]{\includegraphics[width=0.49\linewidth]{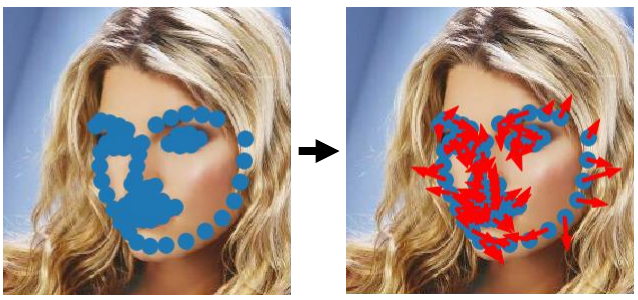}}\hfill
\subfloat[Control Points Estimating]{\includegraphics[width=0.49\linewidth]{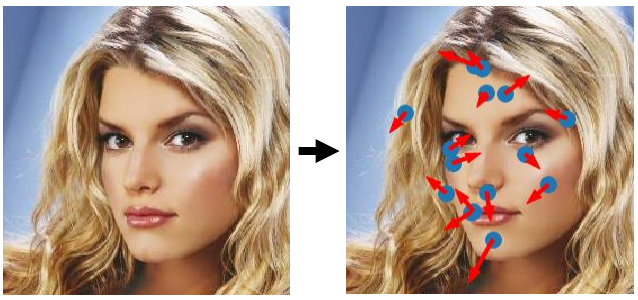}}
\caption{Inputs and outputs of different types of warping modules in neural networks. Given an image, WarpGAN can automatically predict both control points and their displacements based on local features.}
\label{fig:warping}
\end{figure}

\section{Related Work}

\subsection{Automatic Image Warping}
Many works have been proposed to enhance the spatial variability of neural networks via automatic warping. Most of them warp images by predicting a set of global transformation parameters~\cite{jaderberg2015spatial,lin2018stgan} or a dense deformation field~\cite{ganin2016deepwarp}. Parametric methods estimate a small number of global transformation parameters and therefore cannot handle fine-grained local warping while dense deformation needs to predict all the vertices in a deformation grid, most of which are useless and hard to estimate. Cole~\etal~\cite{cole2017synthesizing} first proposed to use spline interpolation in neural networks to allow control point-based warping, but their method requires pre-detected landmarks as input. Several recent works have attempted to combine image warping with GANs to improve the spatial variability of the generator, however these methods either train the warping module separately~\cite{dong2018softgated,carigans}, or need paired data as supervision~\cite{dong2018softgated,siarohin2018deformable_gan}. In comparison, our warping module can be inserted as an enhancement of a normal generator and can be trained as part of an end-to-end system without further modification. To the best of our knowledge, this study is the first work on automatic image warping with self-predicted control points using deep neural networks. An overview of different warping methods are shown in Figure~\ref{fig:warping}.

\subsection{Style Transfer Networks}
Stylizing images by transferring art characteristics has been extensively studied in literature. Given the effective ability of CNNs to extract semantic features~\cite{cnn_1_style_transfer,cnn_2_style_transfer,cnn_3_style_transfer,image_superresolution_style_transfer}, powerful style transfer networks have been developed. Gatys~\etal~\cite{gatys_style_transfer} first proposed a neural style transfer method that uses a CNN to transfer the style content from the style image to the content image. A limitation of this method is that both the style and content images are required to be similar in nature which is not the case for caricatures. Using Generative Adversarial Networks (GANs)~\cite{gan1,gan2} for image synthesis has been a promising field of study, where state-of-the-art results have been demonstrated in applications ranging from text to image translation~\cite{text_to_image_style_transfer}, image inpainting~\cite{image_inpainting_style_transfer}, to image super-resolution~\cite{image_superresolution_style_transfer}. Domain Transfer Network~\cite{taigman_style_transfer}, CycleGAN~\cite{cycle_gan_style_transfer}, StarGAN~\cite{star_gan}, UNIT~\cite{unit}, and MUNIT~\cite{MUNIT} attempt image translation with unpaired image sets. All of these methods only use a de-convolutional network to construct images from the latent space and perform poorly on caricature generation due to the large spatial variation~\cite{carigan,carigans}.

\begin{figure*}[!t]
    \centering
    \captionsetup{font=small}
    \includegraphics[width=\linewidth]{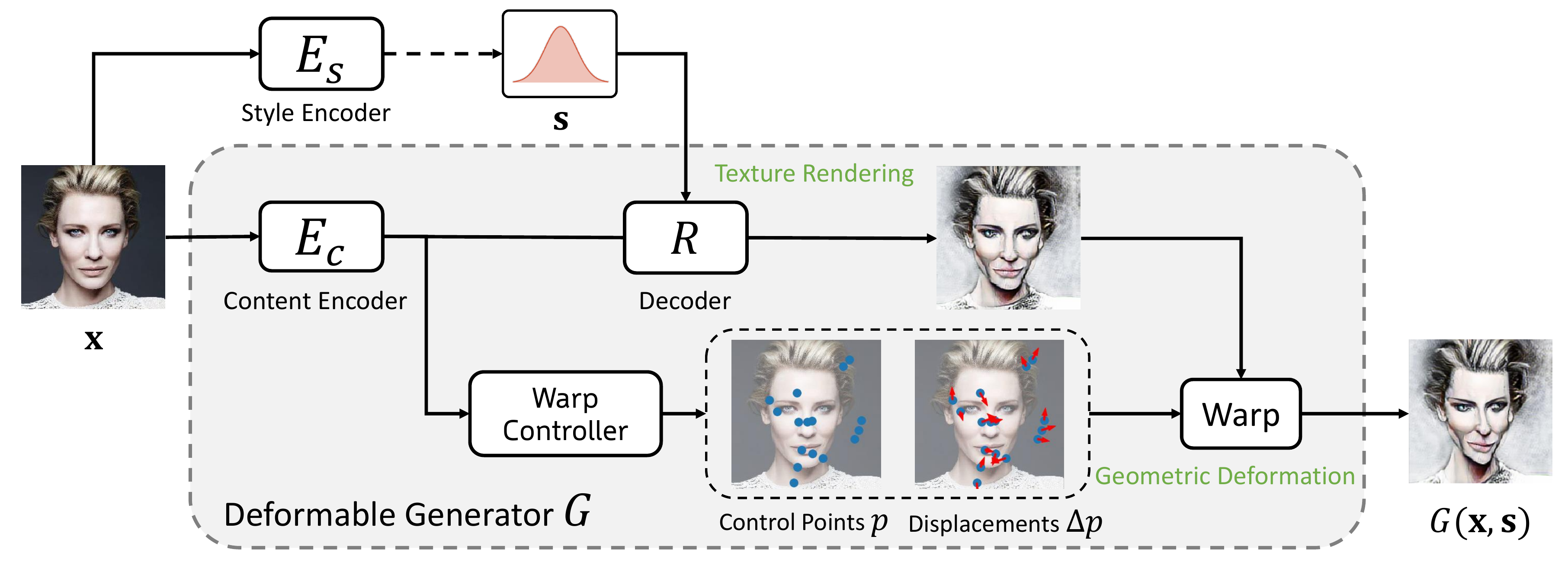}
    \caption{The generator module of WarpGAN. Given a face image, the generator outputs an image with a different texture style and a set of control points along with their displacements. A differentiable module takes the control points and warps the transferred image to generate a caricature.}
    \label{fig:generator}
\end{figure*}

\subsection{Caricature Generation}
Studies on caricature generation can be mainly classified into three categories: deformation-based, texture-based and methods with both. Traditional works mainly focused on exaggerating face shapes by enlarging the deviation of the given shape representation from average, such as 2D landmarks or 3D meshes~\cite{brennan1985caricature,liang2002example,lewiner2011interactive,han2018caricatureshop}, whose deformation capability is usually limited as shape modeling can only happen in the representation space. Recently, with the success of GANs, a few works have attempted to apply style transfer networks to image-to-image caricature generation~\cite{zheng2017photo,carigan}. However, their results suffer from poor visual quality because these networks are not suitable for problems with large spatial variation. Cao et al.~\cite{carigans} recently proposed to decouple texture rendering and geometric deformation with two CycleGANs trained on image and landmark space, respectively. But with their face shape modeled in the PCA subspace of landmarks, they suffer from the same problem of the traditional deformation-based methods. In this work, we propose an end-to-end system with a joint learning of texture rendering and geometric warping. Compared with previous works, WarpGAN can model both shapes and textures in the image space with flexible spatial variability, leading to better visual quality and more artistic shape exaggeration. The differences between caricature generation methods are summarized in Table~\ref{tab:comparison_works}.

\section{Methodology}

\begin{figure}
    \centering
    \captionsetup{font=small}
    \includegraphics[width=\linewidth]{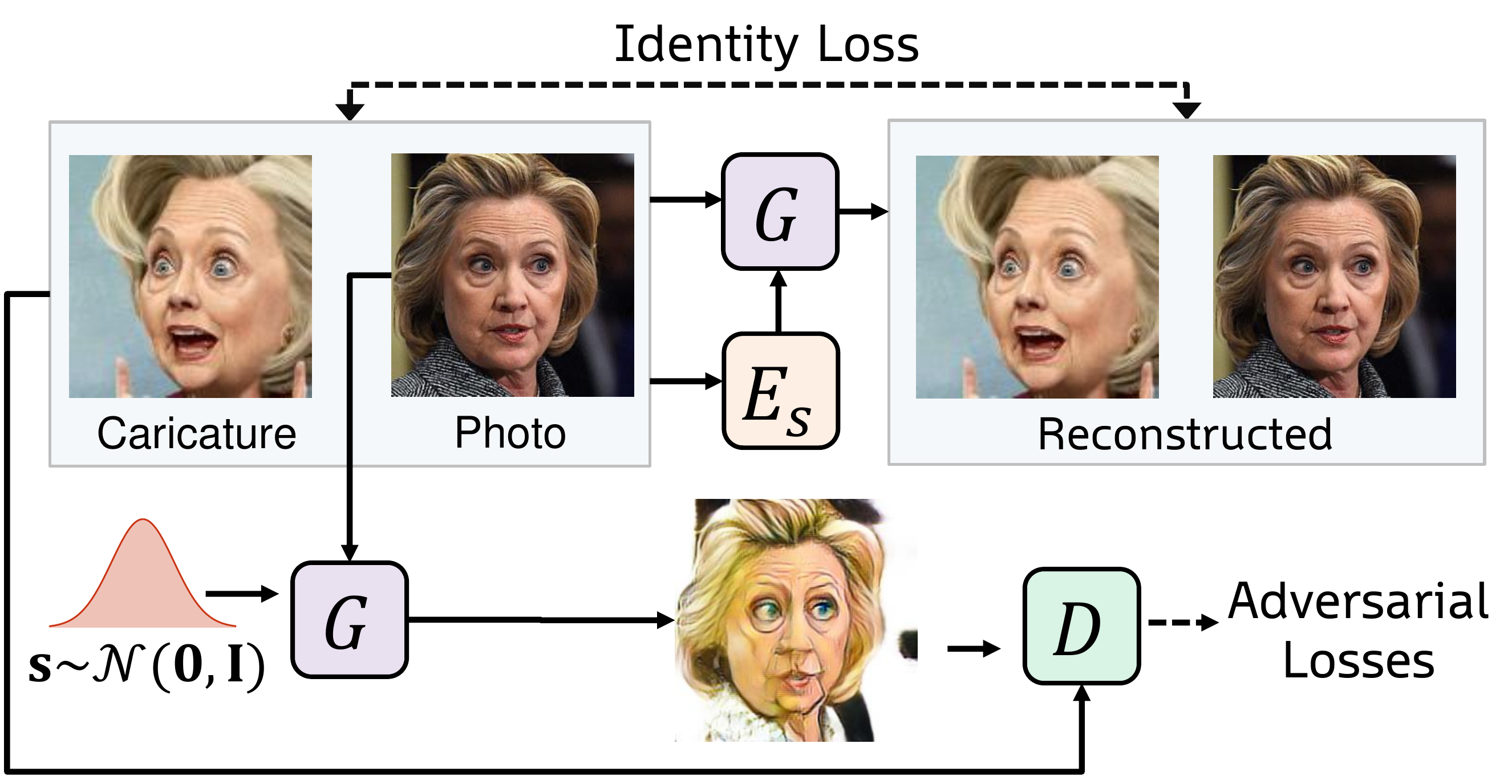}
    \caption{Overview of the proposed WarpGAN.}
    \label{fig:overview}
\end{figure}

Let $\bxp\in \mathcal{X}_p$ be images from the domain of face photos, $\bxc\in \mathcal{X}_c$ be images from the caricature domain and $\bs\in \mathcal{S}$ be the latent codes of texture styles. We aim to build a network that transforms a photo image into a caricature by both transferring its texture style and exaggerating its geometric shape. Our system includes one deformable generator (see Figure~\ref{fig:generator}) $G$, one style encoder $E_s$ and one discriminator $D$ (see Figure~\ref{fig:overview}). The important notations used in this paper are summarized in Table~\ref{tab:notations}.

\begin{table}[!t]
\small
    \captionsetup{font=small}
    \centering
    \begin{tabularx}{\linewidth}{Xc|Xc}
    \toprule
        Name & Meaning & Name & Meaning\\
    \midrule
        $\bxp$ & real photo image &  $y^p$ & label of photo image \\
        $\bxc$ & real caricature image &  $y^c$ & label of caricature image \\
        $E_c$ & content encoder & $R$ & decoder \\
        $E_s$ & style encoder & $D$ & discriminator \\
        $p$ & estimated control points & $\Delta p$ & displacements of $p$ \\
        $M$ & number of identities & $k$ & number of control points\\
    \bottomrule
    \end{tabularx}
    \caption{Important notations used in this paper.}
    \label{tab:notations}
\end{table}

\subsection{Generator}
The proposed deformable generator in WarpGAN is composed of three sub-networks: a content encoder $E_c$, a decoder $R$ and a warp controller. Given any image $\bx\in \IR^{H\times W\times C}$, the encoder outputs a feature map $E_c(\bx)$. Here $H$, $W$ and $C$ are height, width and number of channels respectively. The content decoder takes $E_c(\bx)$ and a random latent style code $\bs\sim \mathcal{N}(0,\mathbf{I})$ to render the given image into an image $R(E_c(\bx),\bs)$ of a certain style. The warp controller estimates the control points and their displacements to warp the rendered images. An overview of the deformable generator is shown in Figure~\ref{fig:generator}.

\begin{figure*}[!t]
    \centering
    \captionsetup{font=small}
    \includegraphics[width=\linewidth]{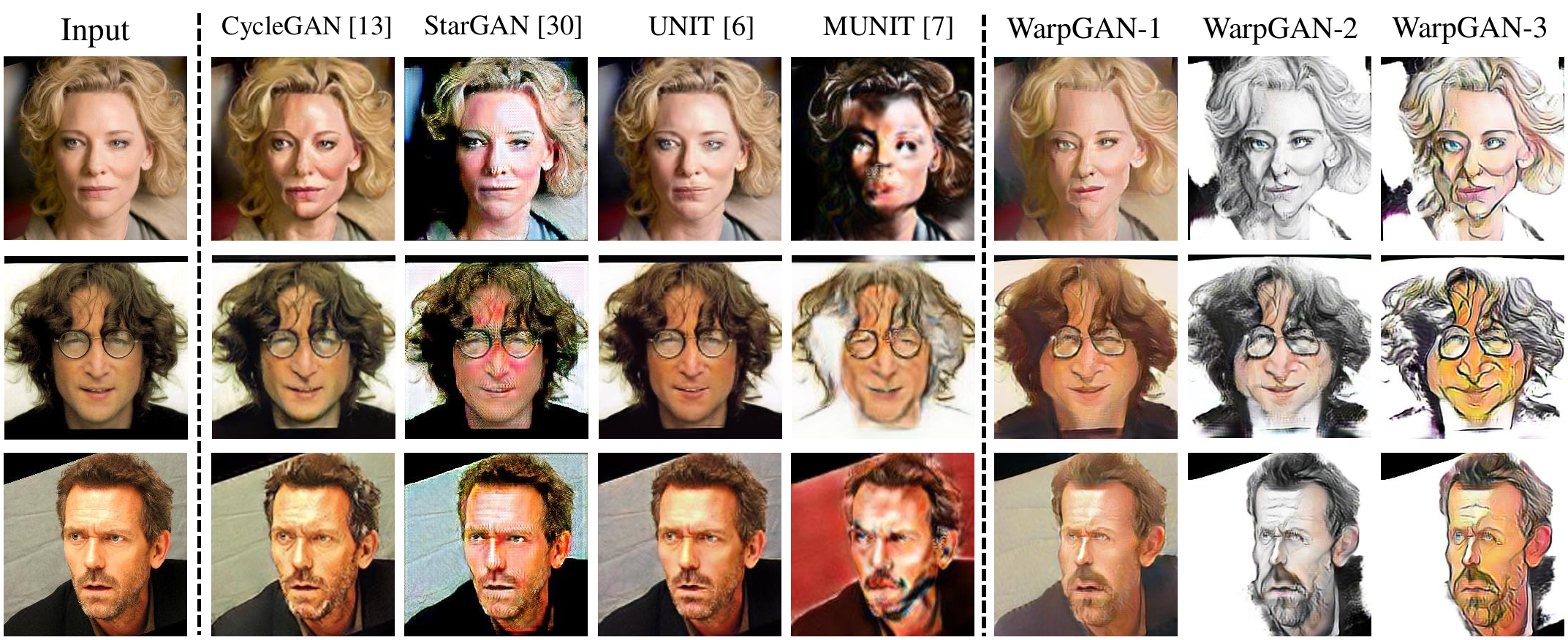}
    \caption{Comparison of 3 different caricature styles from WarpGAN and four other state-of-the-art style transfer networks. WarpGAN is able to deform the faces unlike the baselines.}
    \label{fig:style_baselines}
\end{figure*}

\paragraph{Texture Style Transfer}
Since there is a large variation in the texture styles of caricatures images (See Figure~\ref{fig:frontpage}), we adopt an unsupervised method~\cite{MUNIT} to disentangle the style representation from the feature map $E_c(\bx)$ so that we can transfer the input photo into different texture styles present in the caricature domain. During the training, the latent style code $s\sim \mathcal{N}(0,\mathbf{I})$ is sampled randomly from a normal distribution and passed as an input into the decoder $R$. A multi-layer perceptron in $R$ decodes $\bs$ to generate the parameters of the Adaptive Instance Normalization (AdaIN) layers in $R$, which have been shown to be effective in controlling visual styles~\cite{huang2017adain}. The generated images $R(E_c(\bx),\bs)$ with random styles are then warped and passed to the discriminator. Various styles obtained from WarpGAN can be seen in Figure~\ref{fig:style_baselines}.

To prevent $E_c$ and $R$ from losing semantic information during texture rendering, we combine the identity mapping loss~\cite{taigman_style_transfer} and reconstruction loss~\cite{MUNIT} to regularize $E_c$ and $R$. In particular, a style encoder $E_s$ is used to learn the mapping from the image space to the style space $S$. Given its own style code, both photos and caricatures should be reconstructed from the latent feature map:\vspace{-0.1em}
\begin{align}
    \EL_{idt}^{p} &= \IE_{\bxp\in\mathcal{X}_p}{[\norm{R(E_c(\bxp),E_s(\bxp))-\bxp}_1]} \\
    \EL_{idt}^{c} &= \IE_{\bxc\in\mathcal{X}_c}{[\norm{R(E_c(\bxc),E_s(\bxc))-\bxc}_1]}
\end{align}

\paragraph{Automatic Image Warping}
{
\newcommand{\bpi}{\mathbf{p_i}}
\newcommand{\bq}{\mathbf{q}}
\newcommand{\bp}{\mathbf{p}}
\newcommand{\bw}{\mathbf{w}}
\newcommand{\bv}{\mathbf{v}}
\newcommand{\bb}{\mathbf{b}}

The warp controller is a sub-network of two fully connected layers. With latent feature map $E_c(\bx)$ as input, the controller learns to estimate $k$ control points $p=\{\bp_1,\bp_2,....,\bp_k\}$ and their displacement vectors $\Delta p=\{\Delta \bp_1,\Delta \bp_2, ...\Delta \bp_k\}$, where each $\bp_i$ and $\Delta \bp_i$ is a 2D vector in the u-v space. The points are then fed into a differentiable warping module~\cite{cole2017synthesizing}. Let $p'=\{\bp'_1,\bp'_2,...,\bp'_k\}$ be the destination points, where $\bp'_i=\bp_i+\Delta \bp_i$. A grid sampler of size $H\times W$ can then be computed via thin-plate spline interpolation:\vspace{-0.1em}
\begin{equation}
    f(\bq)=\sum_{i=1}^{k}w_i\phi(||\bq-\bp_i'||)+\bv^T\bq+\bb
\end{equation}
where the vector $\bq$ denotes the u-v location of a pixel in the target image, and $f(\bq)$ gives the inverse mapping of the pixel $\bq$ in the original image, and $\phi(r)=r^2log(r)$ is the kernel function. The parameters $\bw,\bv,\bb$ are fitted to minimize $\sum_{j}^{k}\norm{f(\bp'_j)-\bp_j}^2$ and a curvature constraint, which can be solved in closed form~\cite{glasbey1998review}. With the grid sampler constructed via inverse mapping function $f(\bq)$, the warped image 
\begin{equation}
G(\bx,\bs)=\Warp{(R(E_c(\bx),\bs), p, \Delta p)}
\end{equation} 
can then be generated through bi-linear sampling~\cite{jaderberg2015spatial}. The entire warping module is differentiable and can be trained as part of an end-to-end system.
}

\subsection{Discriminator}
\paragraph{Patch Adversarial Loss}
We first used a fully convolutional network as a patch discriminator~\cite{MUNIT,cycle_gan_style_transfer}. The patch discriminator is trained as a $3$-class classifier to enlarge the difference between the styles of generated images and real photos~\cite{taigman_style_transfer}. Let $D_1$, $D_2$ and $D_3$ denote the logits for the three classes of caricatures, photos and generated images, respectively. The patch adversarial loss is as follows:\vspace{-0.1em}
\begin{align}
    \EL^G_{p} = & -\IE_{\bxp\in \mathcal{X}_p, \bs\in S}{[\log D_1(G(\bxp,\bs))]}\\
\begin{split}
    \EL^D_{p} = & -\IE_{\bxc\in \mathcal{X}_c}{[\log D_1(\bxc)]} - \IE_{\bxp\in \mathcal{X}_p}{[\log D_2(\bxp)]}\\
                    & -\IE_{\bxp\in \mathcal{X}_p, \bs\in S}{[\log D_3(G(\bxp,\bs))]} 
\end{split}
\end{align}

\paragraph{Identity-Preservation Adversarial Loss}
Although patch discriminator is suitable for learning visual style transfer, it fails to capture the distinguishing features of different identities. The exaggeration styles for different people could actually be different based on their facial features (See Section~\ref{sec:shape_style}).
To combine the identity-preservation and identity-specific style learning, we propose to train the discriminator as a $3M$-class classifier, where $M$ is the number of identities. The first, second, and third $M$ classes correspond to different identities of real photos, real caricatures and fake caricatures, respectively. Let $y^p, y^c\in\{1,2,3,...M\}$ be the identity labels of the photos and caricatures, respectively. The identity-preservation adversarial losses for $G$ and $D$ are as follows:\vspace{-0.1em}
\begin{align}
    \EL^G_{g} = & -\IE_{\bxp\in \mathcal{X}_p, \bs\in S}{[\log D(y_p;G(\bxp,\bs))]}\\
\begin{split}
    \EL^D_{g} = & -\IE_{\bxc\in \mathcal{X}_c}{[\log D(y_c;\bxc)]} \\
                    & -\IE_{\bxp\in \mathcal{X}_p}{[\log D(y_p+M;\bxp)]} \\
                    & -\IE_{\bxp\in \mathcal{X}_p, s\in S}{[\log D(y_p+2M;G(\bxp,\bs))]} 
\end{split}
\end{align}

Here, $D(y;x)$ denotes the logits of class $y$ given an image $x$. The discriminator is trained to tell the differences between real photos, real caricatures, generated caricatures as well as the identities in the image. The generator is trained to fool the discriminator in recognizing the generated image as a real caricature of the corresponding identity. 
Finally, the system is optimized in an end-to-end way with the following objective functions:\vspace{-0.1em}
\begin{align}
    & \min_{G} \EL_{G} = \lambda_{p}\EL_{p}^{G} + \lambda_{g}\EL_{g}^{G} + \lambda_{idt}(\EL_{idt}^c+\EL_{idt}^p) \\
    & \min_{D} \EL_{D} = \lambda_{p}\EL_{p}^{D} + \lambda_{g}\EL_{g}^{D}
\end{align}

\begin{figure}
\setlength\tabcolsep{0px}
\newcolumntype{Y}{>{\centering\arraybackslash}X}
    \centering
    \captionsetup{font=small}
    \begin{tabularx}{\linewidth}{YYYYYY}
    \toprule
        Input & w/o $\EL_g$  & w/o $\EL_p$ & w/o $\EL_{idt}$ & with all  \\
    \midrule
        \includegraphics[width=0.97\linewidth]{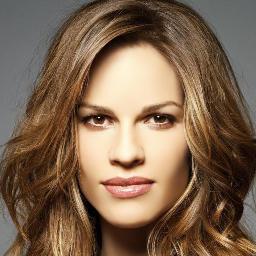} & 
        \includegraphics[width=0.97\linewidth]{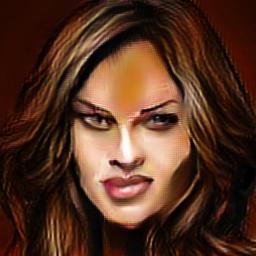} &
        \includegraphics[width=0.97\linewidth]{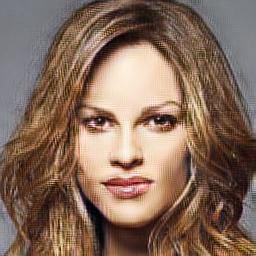} &
        \includegraphics[width=0.97\linewidth]{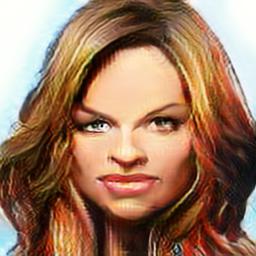} &
        \includegraphics[width=0.97\linewidth]{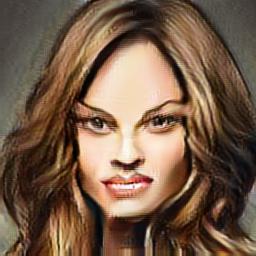} \\
        \includegraphics[width=0.97\linewidth]{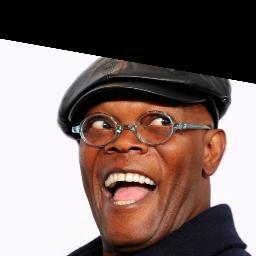} & 
        \includegraphics[width=0.97\linewidth]{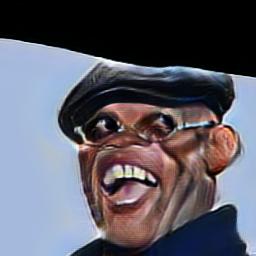} &
        \includegraphics[width=0.97\linewidth]{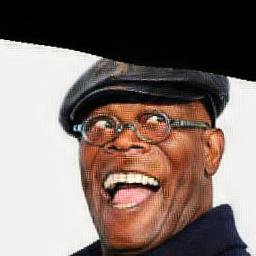} &
        \includegraphics[width=0.97\linewidth]{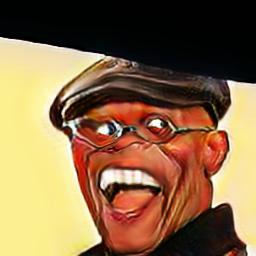} &
        \includegraphics[width=0.97\linewidth]{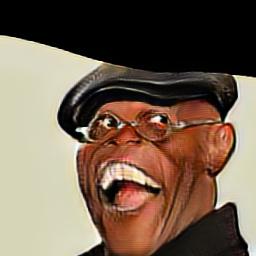} \\
        \includegraphics[width=0.97\linewidth]{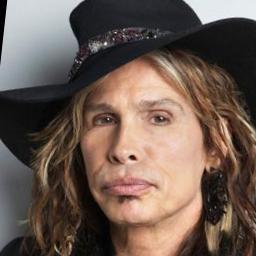} & 
        \includegraphics[width=0.97\linewidth]{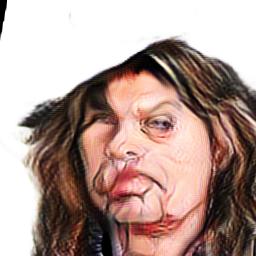} &
        \includegraphics[width=0.97\linewidth]{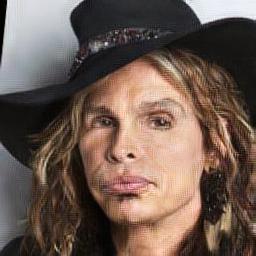} &
        \includegraphics[width=0.97\linewidth]{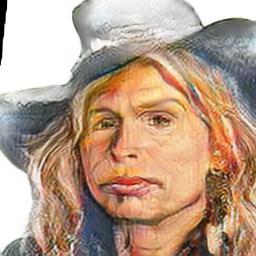} &
        \includegraphics[width=0.97\linewidth]{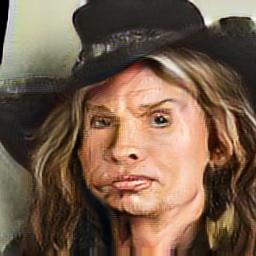} \\
    \bottomrule
    \end{tabularx}
    \caption{Different variants of the WarpGAN without certain loss functions.}
    \label{fig:ablation}
\end{figure}

\section{Experiments}

\begin{figure*}[t]
\footnotesize
\captionsetup{font=small}
\setlength\tabcolsep{1px}
\def\arraystretch{1.1}
\newcolumntype{Y}{>{\centering\arraybackslash}X}
\newcolumntype{V}{>{\centering\arraybackslash}m{1in}}
    \centering
    \begin{tabularx}{\linewidth}{cYYYYYYY}
        & Bigger Eyes &  Smaller Eyes & Longer Face & Shorter Face & Bigger Mouth & Bigger Chin & Bigger Forehead\\
      \noalign{\hrule height 1.2pt}\\[-1.0em]
        \raisebox{3.5\height}{Hand-drawn} & \includegraphics[width=0.99\linewidth]{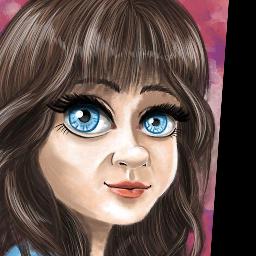} &
        \includegraphics[width=0.99\linewidth]{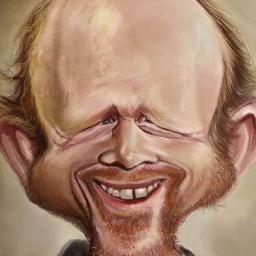} &
        \includegraphics[width=0.99\linewidth]{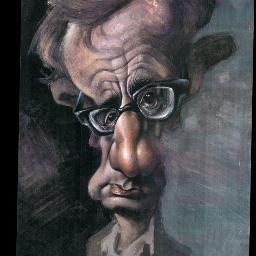} & 
        \includegraphics[width=0.99\linewidth]{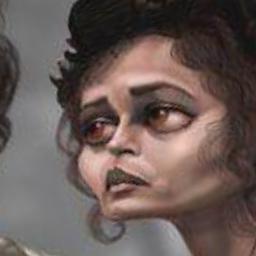} &
        \includegraphics[width=0.99\linewidth]{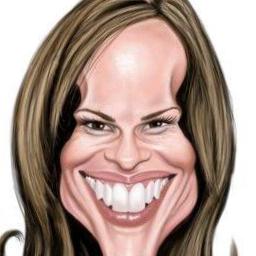} &
        \includegraphics[width=0.99\linewidth]{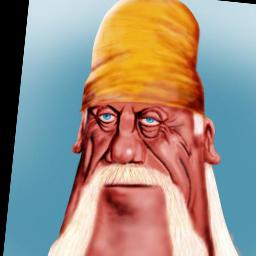} &
        \includegraphics[width=0.99\linewidth]{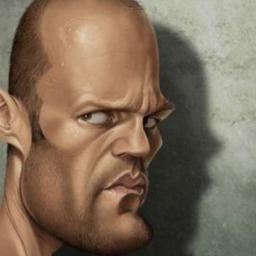} \\
        \raisebox{3.5\height}{WarpGAN Input} & \includegraphics[width=0.99\linewidth]{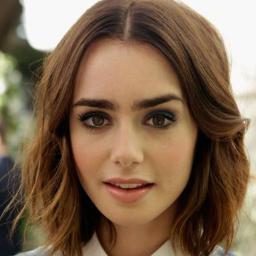} &
        \includegraphics[width=0.99\linewidth]{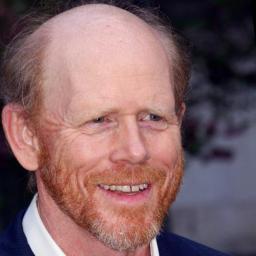} &
        \includegraphics[width=0.99\linewidth]{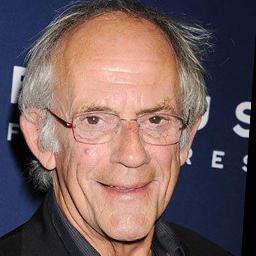} & 
        \includegraphics[width=0.99\linewidth]{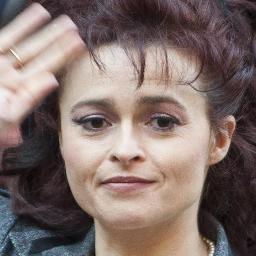} &
        \includegraphics[width=0.99\linewidth]{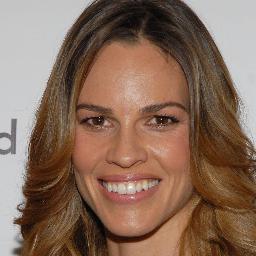} &
        \includegraphics[width=0.99\linewidth]{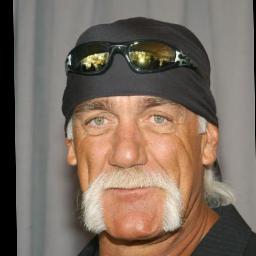} &
        \includegraphics[width=0.99\linewidth]{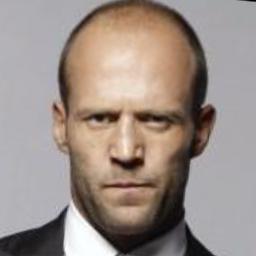} \\
        \raisebox{3.5\height}{WarpGAN Output} & \includegraphics[width=0.99\linewidth]{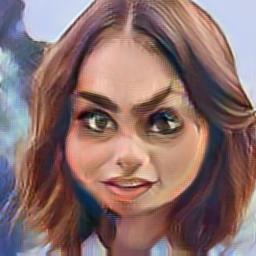} &
        \includegraphics[width=0.99\linewidth]{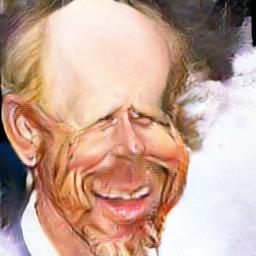} &
        \includegraphics[width=0.99\linewidth]{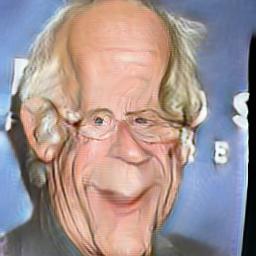} & 
        \includegraphics[width=0.99\linewidth]{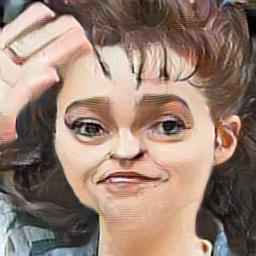} &
        \includegraphics[width=0.99\linewidth]{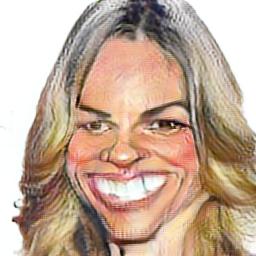} &
        \includegraphics[width=0.99\linewidth]{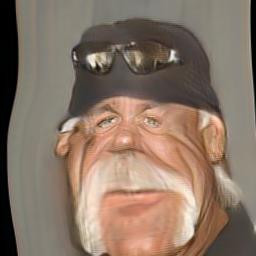} &
        \includegraphics[width=0.99\linewidth]{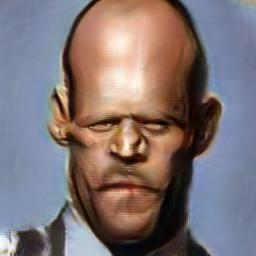} \\
    \noalign{\hrule height 1.0pt}
    \end{tabularx}
    \caption{A few typical exaggeration styles learned by WarpGAN. First row shows hand-drawn caricatures that have certain exaggeration styles. The second and third row show the input images and the generated images of WarpGAN with the corresponding exaggeration styles. All the identities are from the testing set.}
    \label{fig:modes}
\end{figure*}

\paragraph{Dataset} We use the images from a public domain dataset,  WebCaricature~\cite{huo2017webcaricature}\footnote{\url{https://cs.nju.edu.cn/rl/WebCaricature.htm}}, to conduct the experiments. The dataset consists of $6,042$ caricatures and $5,974$ photos from $252$ identities. We align all the images with five landmarks. Then, the images are aligned through similarity transformation using the five landmarks and are resized to $256\times256$. We randomly split the dataset into a training set of $126$ identities ($3,016$ photos and $3,112$ caricatures) and a testing set of $126$ identities ($2,958$ photos and $2,930$ caricatures). \textit{All the testing images in this paper are from identities in the testing set}.

\paragraph{Training Details} We use ADAM optimizers in Tensorflow with $\beta_1=0.5$ and $\beta_2=0.9$ for the whole network. Each mini-batch consists of a random pair of photo and caricature. We train the network for $100,000$ steps. The learning rate starts with $0.0001$ and is decreased linearly to $0$ after $50,000$ steps. We empirically set $\lambda_{g}=1.0$, $\lambda_{p}=2.0$, $\lambda_{idt}=10.0$ and number of control points $k=16$. We conduct all experiments using Tensorflow r1.9 and one Geforce GTX 1080 Ti GPU. The average speed for generating one caricature image on this GPU is $0.082$s. The details of the architecture are provided in the supplementary material.

% \paragraph{Dataset} We use the images from a public domain dataset,  WebCaricature~\cite{huo2017webcaricature}\footnote{\url{https://cs.nju.edu.cn/rl/WebCaricature.htm}}, to conduct the experiments. The dataset consists of $6,042$ caricatures and $5,974$ photos from $252$ identities. We align all the images via similarity transformation and resize them to $256\times256$.
% We randomly split the dataset into a training set of $126$ identities ($3,016$ photos and $3,112$ caricatures) and a testing set of $126$ identities ($2,958$ photos and $2,930$ caricatures). \textit{All of the images shown in this section are from the testing set}.

% \paragraph{Training Details} We use ADAM optimizers in Tensorflow with $\beta_1=0.5$ and $\beta_2=0.9$ for the whole network. We train the network for $100,000$ steps. The learning rate starts with $0.0001$ and is decreased linearly to $0$ after $50,000$ steps. We empirically set $\lambda_{g}=1.0$, $\lambda_{p}=2.0$, $\lambda_{idt}=10.0$ and number of control points $k=16$. Due to space limitations, we provide the details of the architecture in the Appendix.

\subsection{Comparison to State-of-the-Art}
We qualitatively compare our caricature generation method with \textbf{CycleGAN}~\cite{cycle_gan_style_transfer}, \textbf{StarGAN}~\cite{star_gan}, ~\textbf{Unsupervised Image-to-Image Translation (UNIT)}~\cite{unit}, and Multimodal UNsupervised Image-to-image Translation (\textbf{MUNIT})~\cite{MUNIT} for style transfer approaches\footnote{We train the baselines using their official implementations.}. We find that among all the three baseline style transfer networks, CycleGAN and MUNIT demonstrate the most visually appealing texture styles (see Figure~\ref{fig:style_baselines}).  StarGAN and UNIT produce very photo-like images with minimal or erroneous changes in texture. Since all these networks focus only on transferring the texture styles, they fail to deform the faces into caricatures, unlike WarpGAN. The other issue with the baselines methods is that they do not have a module for warping the images and therefore, they try to compensate for deformations in the face using only texture. Due to the complexity of this task, it becomes increasingly difficult to train them and they usually result in generating collapsed images.

\subsection{Ablation Study}
\label{sec:ablation}
% To analyze the function of different modules in our system, we train three variants of WarpGAN for comparison by removing $\EL_{g}$, $\EL_{p}$ and $\EL_{adv}$, respectively. Figure~\ref{fig:ablation} shows a comparison of WarpGAN variants that include all the loss functions. Without the proposed identity-preservation adversarial loss, the discriminator in Model A only focuses on local texture styles and therefore the geometric warping fails to capture personal features and is close to randomness. Without the patch adversarial loss, the discriminator in ''Model B'' mainly focuses on facial shape and the model fails to learn diverse texture styles. The ``Model C'', without identity mapping loss, still performs well in terms of texture rendering and shape exaggeration. However, the style is not fully disentangled and cannot be controlled by the latent style code.
To analyze the function of different modules in our system, we train three variants of WarpGAN for comparison by removing $\EL_{g}$, $\EL_{p}$ and $\EL_{idt}$, respectively. Figure~\ref{fig:ablation} shows a comparison of WarpGAN variants that include all the loss functions. Without the proposed identity-preservation adversarial loss, the discriminator only focuses on local texture styles and therefore the geometric warping fails to capture personal features and is close to randomness. Without the patch adversarial loss, the discriminator mainly focuses on facial shape and the model fails to learn diverse texture styles. The model without identity mapping loss still performs well in terms of texture rendering and shape exaggeration. We keep the identity loss to improve the visual quality of the generated images.

\begin{figure}[t]
\setlength\tabcolsep{0px}
\newcolumntype{Y}{>{\centering\arraybackslash}X}
    \centering
    \captionsetup{font=small}
    \begin{tabularx}{\linewidth}{YYYYY}
    \toprule
        Input & $\alpha=0.5$ & $\alpha=1.0$ & $\alpha=1.5$ & $\alpha=2.0$ \\
    \midrule
        \includegraphics[width=0.97\linewidth]{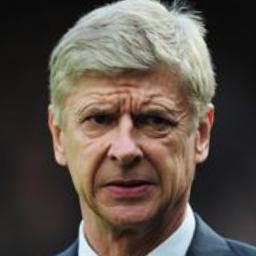} & 
        \includegraphics[width=0.97\linewidth]{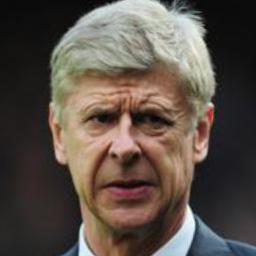} & 
        \includegraphics[width=0.97\linewidth]{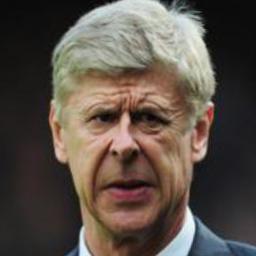} & 
        \includegraphics[width=0.97\linewidth]{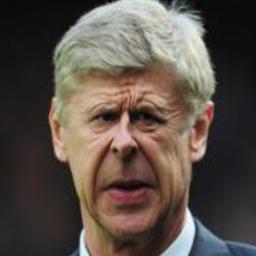} & 
        \includegraphics[width=0.97\linewidth]{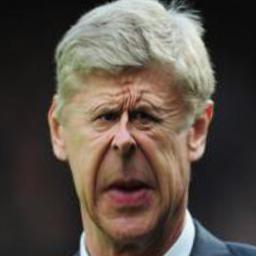} \\
        \includegraphics[width=0.97\linewidth]{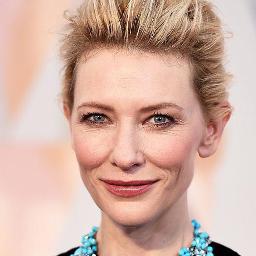} & 
        \includegraphics[width=0.97\linewidth]{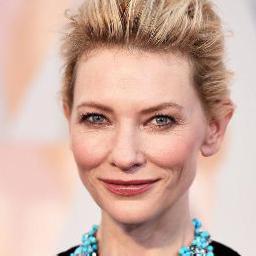} & 
        \includegraphics[width=0.97\linewidth]{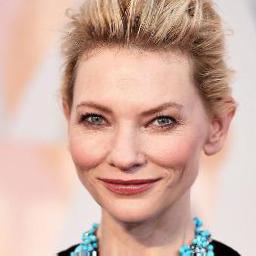} & 
        \includegraphics[width=0.97\linewidth]{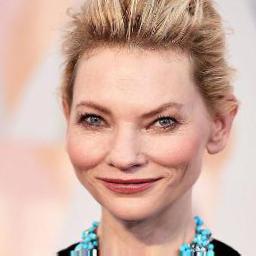} & 
        \includegraphics[width=0.97\linewidth]{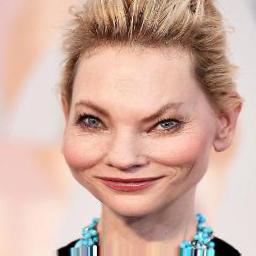} \\
    \bottomrule
    \end{tabularx}
    \caption{The result of changing the amount of exaggeration by scaling the $\Delta p$ with an input parameter $\alpha$. }
    \label{fig:scales}
\end{figure}

\subsection{Shape Exaggeration Styles}
\label{sec:shape_style}
Caricaturists usually define a set of prototypes of face parts and have certain modes on how to exaggerate them~\cite{klare2012towards}.
In WarpGAN we do not adopt any method to exaggerate the facial regions explicitly, but instead we introduce the identity preservation constraint as part of the adversarial loss. This forces the network to exaggerate the faces to be more distinctive from other identities and implicitly encourages the network to learn different exaggeration styles for people with different salient features. Some example exaggeration styles learned by the network are shown in Figure~\ref{fig:modes}. 

\subsection{Customizing the exaggeration}
Although the WarpGAN is trained as a deterministic model, we introduce a parameter $\alpha$ during deployment to allow customization of the exaggeration extent. Before warping, the displacement of control points $\Delta p$ will be scaled by $\alpha$ to control how much the face shape will be exaggerated. The results are shown in Figure~\ref{fig:scales}. When $\alpha=0.0$, only the texture is changed and $\alpha=1.0$ leads to the original output of the WarpGAN. Even when changing $\alpha$ to $2.0$, the resulting images appear as caricatures, but only the distinguishing facial features are exaggerated. Since the texture styles are learned in a disentangled way, WarpGAN can generate various texture styles. Figure~\ref{fig:style_baselines} shows results from WarpGAN with three randomly sampled styles.

\subsection{Quantitative Analysis}

\paragraph{Face Recognition} In order to quantify identity preservation accuracy for caricatures generated by WarpGAN, we evaluate automatic face recognition performance using two state-of-the-art face matchers: (1) a Commercial-Off-The-Shelf (COTS) matcher\footnote{Uses a convolutional neural network for face recognition.} and (2) an open source SphereFace~\cite{sphere_face} matcher.
\begin{table}[t]
\centering
\small
\captionsetup{font=small}
\caption{Rank-1 identification accuracy for three different matching protocols using two state-of-the-art face matchers, COTS and SphereFace~\cite{sphere_face}.}
\resizebox{\textwidth}{!}{
\begin{tabular}{|l||c||c|}
\noalign{\hrule height 1.5pt}
                \textbf{Method} & \textbf{COTS}
                & \textbf{SphereFace~\cite{sphere_face}} \\ \hline
\noalign{\hrule height 1.2pt}
Photo-to-Photo &  94.81 $\pm$ 1.22\% & 90.78 $\pm$ 0.64\%\\
Hand-drawn-to-Photo & 41.26 $\pm$ 1.16\% & 45.80 $\pm$ 1.56\%\\
WarpGAN-to-Photo &  79.00 $\pm$ 1.46\% & 72.65 $\pm$ 0.84\%\\
\noalign{\hrule height 1.5pt}
\end{tabular}
}
\label{tab:recognition}
\end{table}

An identification experiment is conducted where one photo of the identity is kept in the gallery while all remaining photos, or all hand-drawn caricatures, or all synthesized caricatures for the same identity are used as probes. We evaluate the Rank-1 identification accuracy using 10-fold cross validation and report the mean and standard deviation across the folds in Table~\ref{tab:recognition}. We find that the generated caricatures can be matched to real face images with a higher accuracy than hand-drawn caricatures. We also observe the same trend for both the matchers, which suggests that recognition on synthesized caricatures is consistent and matcher-independent.

\paragraph{Perceptual Study} We conducted two perceptual studies by recruiting 5 caricature artists who are experts in their field to compare hand-drawn caricatures with images synthesized by our baselines along with our WarpGAN. A caricature is generated from a random image for each 126 subjects in the WebCaricature testing set. The first perceptual study uses $30$ of them and $96$ are used for the second. Experts do not have any knowledge of the source of the caricatures and they rely solely on their perceptual judgment.

The first study assesses the overall similarity of the generated caricatures to the hand-drawn ones. Each caricature expert was shown a face photograph of a subject along with three corresponding caricatures generated by CycleGAN, MUNIT, and WarpGAN, respectively. The experts then rank each of the three generated caricatures from ``most visually closer to a hand-drawn caricature'' to ``least similar to a hand-drawn caricature''. We find that caricatures generated by WarpGAN is ranked as the most similar to a real caricature $99\%$ of the time, compared to $0.5\%$ and $0.5\%$ for CycleGAN and MUNIT, respectively.

\begin{table}[!t]
\centering
\small
\captionsetup{font=small}
\caption{Average perceptual scores from 5 caricature experts for visual quality and exaggeration extent. Scores range from 1 to 10.}
\begin{tabular}{|l|c|c|}
\noalign{\hrule height 1.5pt}
\textbf{Method} & \textbf{Visual Quality} & \textbf{Exaggeration}\\ 
\noalign{\hrule height 1.2pt}
Hand-Drawn & 7.70 & 7.16 \\
CycleGAN~\cite{cycle_gan_style_transfer} & 2.43 & 2.27\\
MUNIT~\cite{MUNIT} & 1.82 & 1.83 \\
\textbf{WarpGAN} & \textbf{5.61} & \textbf{4.87}\\
\noalign{\hrule height 1.5pt}
\end{tabular}
\label{tab:perceptual}
\end{table}

In the second study, experts scored the generated caricatures according to two criteria: (i) visual quality, and (ii) whether the caricatures are exaggerated in proper manner where only prominent facial features are deformed. Experts are shown three photographs of a subject along with a caricature image that can either be (i) a real hand-drawn caricature, or (ii) generated using one of the three automatic style transfer methods. From Table~\ref{tab:perceptual} we find that WarpGAN receives the best perceptual scores out of the three methods. Even though hand-drawn caricatures rate higher, our approach, WarpGAN, has made a tremendous leap in automatically generating caricatures, especially when compared to state-of-the-art.

\begin{figure}[!t]
\setlength\tabcolsep{0px}
\newcolumntype{Y}{>{\centering\arraybackslash}X}
    \centering
    \small
    \captionsetup{font=small}
    \begin{tabularx}{\linewidth}{YYYY}
        Input & Warping Only & Texture Only & Both \\
        \includegraphics[width=0.97\linewidth]{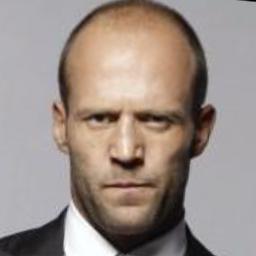} & 
        \includegraphics[width=0.97\linewidth]{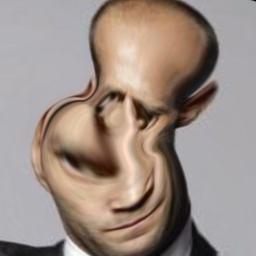} & 
        \includegraphics[width=0.97\linewidth]{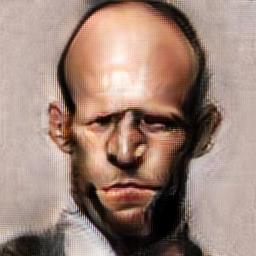} & 
        \includegraphics[width=0.97\linewidth]{fig/modes/big_forehead_4.jpg}
    \end{tabularx}
    \caption{Example result images generated by the WarpGAN trained without texture/warping and with both. }
    \label{fig:texture_shape}
\end{figure}

\section{Discussion}
% In this section, we discuss the three key differences between our work and previous works on GAN-based visual style transfer and caricature generation:
\paragraph{Joint Rendering and Warping Learning} Unlike other visual style transfer tasks~\cite{taigman_style_transfer,cycle_gan_style_transfer,MUNIT}, transforming photos into caricatures involves both texture difference and geometric transition. Texture is import in exaggerating local fine-grained features such as depth of the wrinkles while geometric deformation allows exaggeration of global features such as face shape. Conventional style transfer networks~\cite{taigman_style_transfer,cycle_gan_style_transfer,MUNIT} aims to reconstruct an image from feature space using a decoder network. Because the decoder is a stack of nonlinear local filters, they are intrinsically inflexible in terms of spatial variation and the decoded images usually suffer from poor quality and severe information loss when there is a large geometric discrepancy between the input and output domain. On the other hand, warping-based methods are limited by nature to not being able to change the content and fine-grained details. Therefore, both style transfer and warping module are necessary parts for our adversarial learning framework. As shown in Figure~\ref{fig:ablation}, without either module, the generator will not be able to close the gap between photos and caricatures and the balance of competition between generator and discriminator will be broken, leading to collapsed results. 

\vspace{-0.8em}\paragraph{Identity-preservation Adversarial Loss}
The discriminator in conventional GANs are usually trained as a binary~\cite{cycle_gan_style_transfer} or ternary classifiers~\cite{taigman_style_transfer}, with each class representing a visual style. However, we found that because of the large variation of shape exaggeration in the caricatures, treating all the caricatures as one class in the discriminator would lead to the confusion of the generator, as shown in Figure~\ref{fig:ablation}. However, we observe that caricaturists tend to give similar exaggeration styles to the same person. Therefore, we treat each identity-domain pair as a separate class to reduce the difficulty of learning and also encourage the identity-preservation after the shape exaggeration.

\section{Conclusion}
In this paper, we proposed a new method of caricature generation, namely WarpGAN, that addresses both style transfer and face deformation in a joint learning framework. Without explicitly requiring any facial landmarks, the identity-preserving adversarial loss introduced in this work appropriately learns to capture caricature artists' style while preserving the identity in the generated caricatures. We evaluated the generated caricatures by matching synthesized caricatures to real photos and observed that the recognition accuracy is higher than caricatures drawn by artists. Moreover, five caricature experts suggest that caricatures synthesized by WarpGAN are not only pleasing to the eye, but are also realistic where only the appropriate facial features are exaggerated and that our WarpGAN indeed outperforms the state-of-the-art networks.

\newpage
{\small
\bibliographystyle{unsrt}
\bibliography{egbib}
}

\appendix

\section{Implementation Details}
\paragraph{Preprocessing} 
We align all the images with five landmarks (left eye, right eye, nose, mouth left, mouth right) using the ones provided in the WebCaricature dataset~\cite{huo2017webcaricature} protocol. Since the protocol does not provide the locations of eye centers, we estimate them by taking the average of the corresponding eye corners. Then, a similarity transformation is applied for all the images using the five landmarks. The aligned images are resized to $256\times256$. The whole dataset consists of $6,042$ caricatures and $5,974$ photos from $252$ identities. We randomly split the dataset into a training set of $126$ identities ($3,016$ photos and $3,112$ caricatures) and a testing set of $126$ identities ($2,958$ photos and $2,930$ caricatures). \textbf{All the testing images in the main paper and this supplementary material are from the identities in the testing split}.

\paragraph{Experiment Settings} 
We conduct all experiments using Tensorflow r1.9 and one Geforce GTX 1080 Ti GPU. The average speed for generating one caricature image on this GPU is $0.082$s.

\paragraph{Architecture}
Our network architecture is modified based on MUNIT~\cite{MUNIT}. Let \texttt{c7s1-k} be a $7\times7$ convolutional layer with $k$ filters and stride $1$. \texttt{dk} denotes a $4\times 4$ convolutional layer with $k$ filters and stride $2$. \texttt{Rk} denotes a residual block that contains two $3\times 3$ convolutional layers. \texttt{uk} denotes a $2\times$
upsampling layer followed by a $5\times 5$ convolutional layer with $k$ filters and stride $1$. \texttt{fck} denotes a fully connected layer with $k$ filters. \texttt{avgpool} denotes a global average pooling layer. We apply Instance Normalization (IN)~\cite{ulyanov2017instancenorm}
to the content encoder and Adaptive Instance Normalization (AdaIN)~\cite{huang2017adain} to the decoder. No normalization is used in the style encoder. We use Leaky ReLU with slope 0.2 in the discriminator and ReLU activation everywhere else. The architectures of different modules are as follows:
\begin{itemize}
    \itemsep0em
    \item Style Encoder: \\ \texttt{c7s1-64,d128,d256,avgpool,fc8}
    \item Content Encoder: \\ \texttt{c7s1-64,d128,d256,R256,R256,R256}
    \item Decoder: \\ \texttt{R256,R256,R256,u128,u64,c7s1-3}
    \item Discriminator: \\ \texttt{d32,d64,d128,d256,d512,fc512,fc3M}
\end{itemize}
A separate branch of $1\times 1$ convolutional layer with $3$ filters and stride $1$ is attached to the last convolutional layer of the discriminator to output $D_1,D_2,D_3$ for patch adversarial losses. The style decoder (the multi-layer perceptron) has two hidden fully connected layers of $128$ filters without normalization and the warp controller has only one hidden fully connected layer of $128$ filters with Layer Normalization~\cite{ba2016layernorm}. The length of the latent style code is set to $8$.

\begin{figure*}[t]
\setlength\tabcolsep{5px}
\newcolumntype{Y}{>{\centering\arraybackslash}X}
    \centering
    \begin{tabularx}{\linewidth}{YYY|YYY}
    \toprule
        \includegraphics[width=0.97\linewidth]{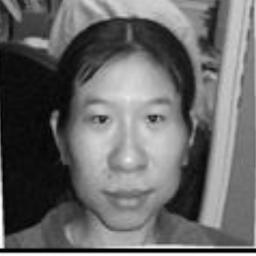} & 
        \includegraphics[width=0.97\linewidth]{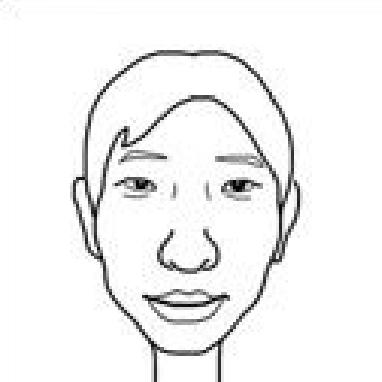} &
        \includegraphics[width=0.97\linewidth]{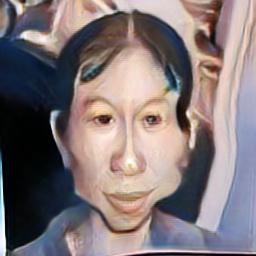} & 
        \includegraphics[width=0.97\linewidth]{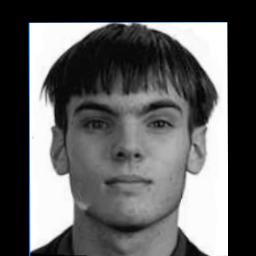} & 
        \includegraphics[width=0.97\linewidth]{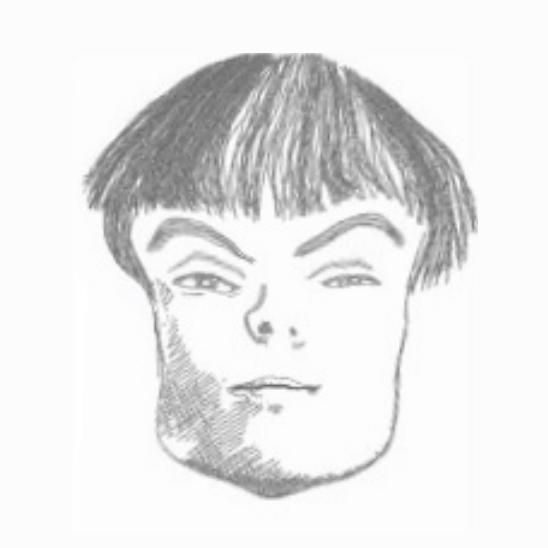} &
        \includegraphics[width=0.97\linewidth]{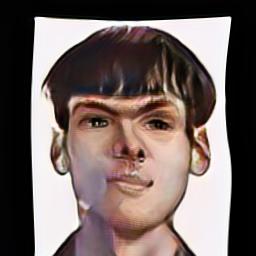} \\
        Input & Liang~\etal\cite{liang2002example} & Ours & Input & Mo~\etal\cite{mo2004improved} & Ours \\\midrule
        % &&&&&\\[-0.8em]
        \includegraphics[width=0.97\linewidth]{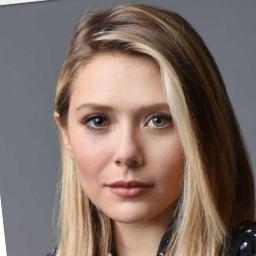} & 
        \includegraphics[width=0.97\linewidth]{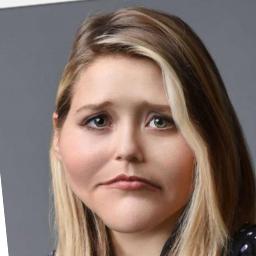} &
        \includegraphics[width=0.97\linewidth]{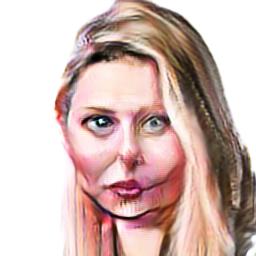} & 
        \includegraphics[width=0.97\linewidth]{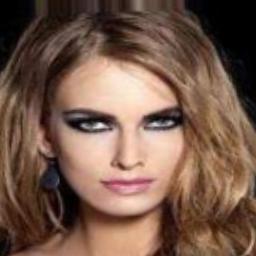} &         
        \includegraphics[width=0.97\linewidth]{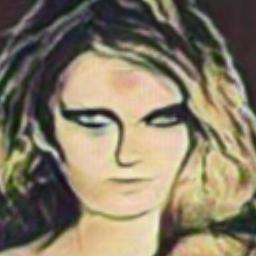} &
        \includegraphics[width=0.97\linewidth]{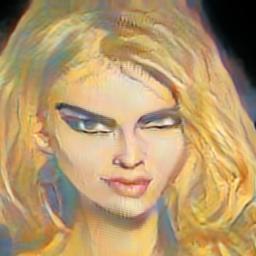} \\
        Input & Han~\etal\cite{han2018caricatureshop} & Ours & Input & Zheng~\etal\cite{zheng2017photo} & Ours \\\midrule
        % &&&&&\\[-0.8em]
        \includegraphics[width=0.97\linewidth]{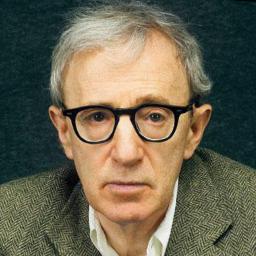} & 
        \includegraphics[width=0.97\linewidth]{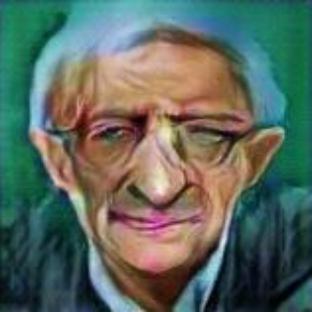} &
        \includegraphics[width=0.97\linewidth]{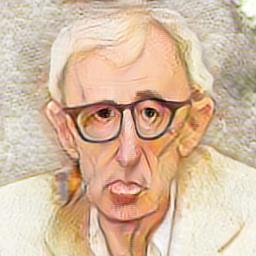} & 
        \includegraphics[width=0.97\linewidth]{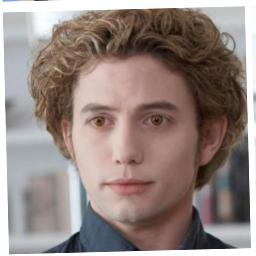} & 
        \includegraphics[width=0.97\linewidth]{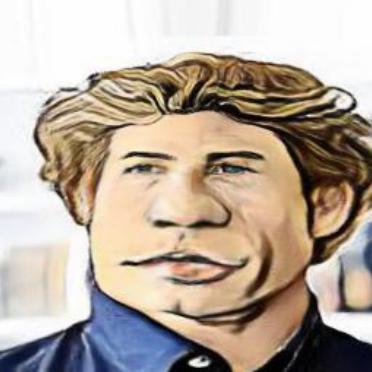} &
        \includegraphics[width=0.97\linewidth]{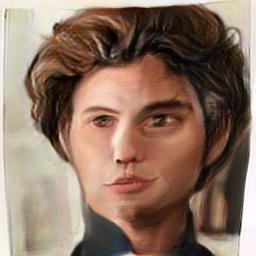} \\
        Input & CariGAN\cite{carigan} & Ours & Input & CariGANs\cite{carigans} & Ours \\
    \bottomrule
    \end{tabularx}
    \caption{Comparison with previous works on caricature generation. In each cell, the left and middle images are the input and result images taken from the baseline paper, respectively. The right images are the results of WarpGAN.}
    \label{appendix:fig:baselines}
\end{figure*}

\section{Additional Baselines}

In the main paper, we compared WarpGAN with state-of-the-art style transfer networks as baselines. Here, we compare WarpGAN with other caricature generation works~\cite{liang2002example,mo2004improved,han2018caricatureshop,zheng2017photo,carigan,carigans}. Since these methods do not release their code and use different testing images, we crop the images from their papers and compare with them one by one. All the baseline results are also taken from their original papers. The results are shown in Figure~\ref{appendix:fig:baselines}.

\section{Transformation Methods}
\begin{figure*}[t]
\setlength\tabcolsep{0px}
\newcolumntype{Y}{>{\centering\arraybackslash}X}
    \centering
    \begin{tabularx}{\linewidth}{YYYYYY}
        & Input & Projective transformation &  Dense deformation  & Landmark-based & Ours \\\midrule
        \raisebox{4em}{Image} &
        \includegraphics[width=0.97\linewidth,trim={33px 24px 11px 10px},clip]{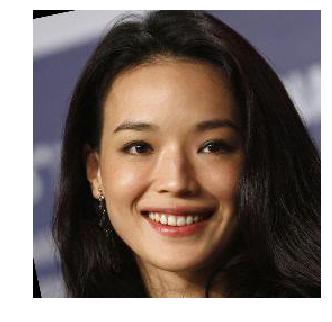} & 
        \includegraphics[width=0.97\linewidth,trim={33px 24px 11px 10px},clip]{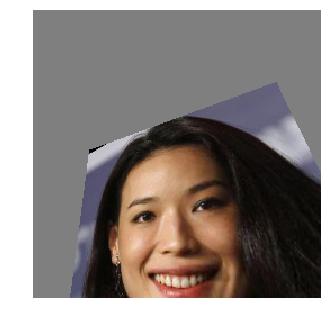} &
        \includegraphics[width=0.97\linewidth,trim={33px 24px 11px 10px},clip]{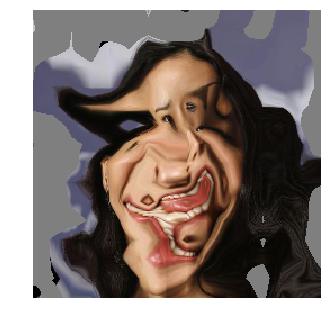} &
        \includegraphics[width=0.97\linewidth,trim={33px 24px 11px 10px},clip]{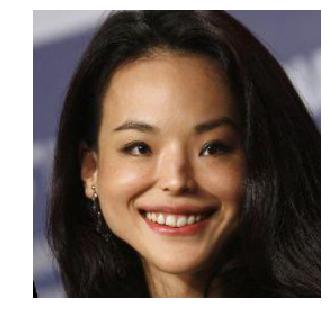} &
        \includegraphics[width=0.97\linewidth,trim={33px 24px 11px 10px},clip]{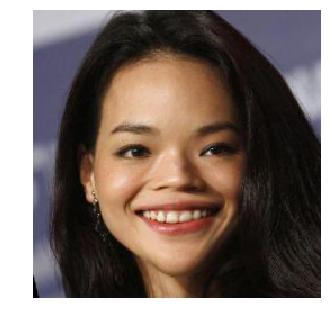} \\
        \raisebox{4em}{Transformation} &
        \includegraphics[width=0.97\linewidth,trim={33px 24px 11px 10px},clip]{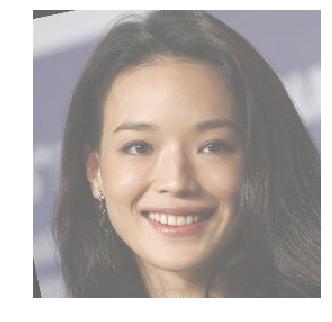} & 
        \includegraphics[width=0.97\linewidth,trim={33px 24px 11px 10px},clip]{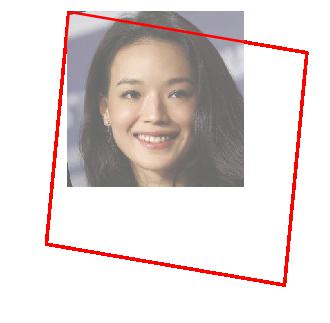} &
        \includegraphics[width=0.97\linewidth,trim={33px 36px 22px 10px},clip]{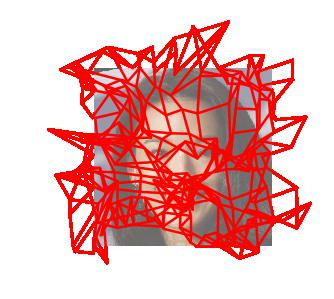} &
        \includegraphics[width=0.97\linewidth,trim={33px 24px 11px 10px},clip]{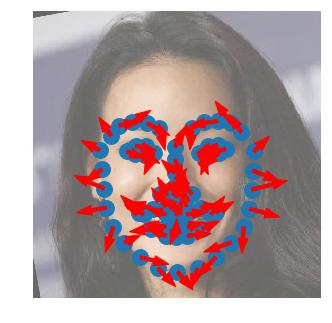} &
        \includegraphics[width=0.97\linewidth,trim={33px 24px 11px 10px},clip]{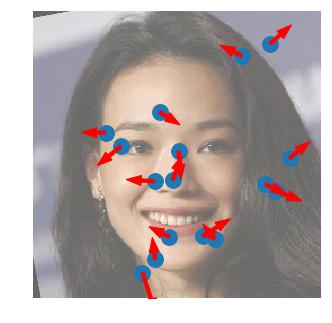} \\\midrule
        \raisebox{4em}{Image} &
        \includegraphics[width=0.97\linewidth,trim={33px 24px 11px 10px},clip]{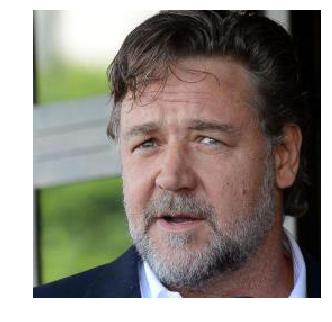} & 
        \includegraphics[width=0.97\linewidth,trim={33px 24px 11px 10px},clip]{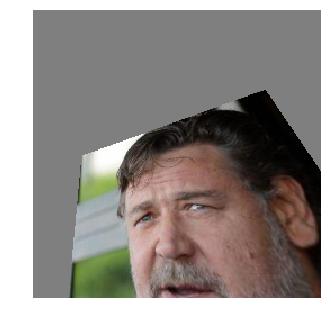} &
        \includegraphics[width=0.97\linewidth,trim={33px 24px 11px 10px},clip]{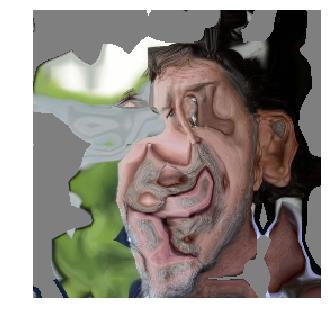} &
        \includegraphics[width=0.97\linewidth,trim={33px 24px 11px 10px},clip]{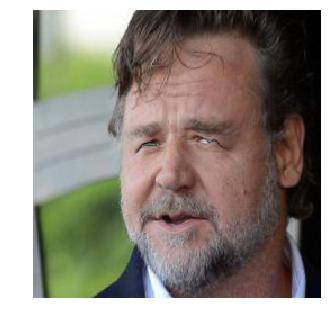} &
        \includegraphics[width=0.97\linewidth,trim={33px 24px 11px 10px},clip]{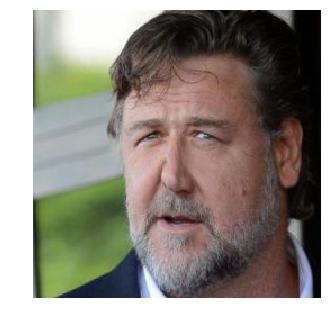} \\
        \raisebox{4em}{Transformation} &
        \includegraphics[width=0.97\linewidth,trim={33px 24px 11px 10px},clip]{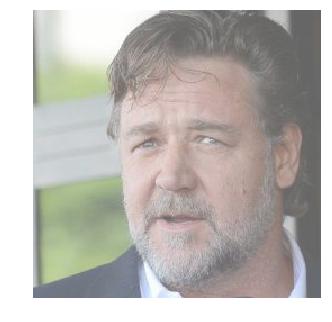} & 
        \includegraphics[width=0.97\linewidth,trim={33px 24px 11px 10px},clip]{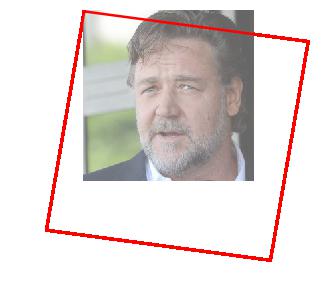} &
        \includegraphics[width=0.97\linewidth,trim={33px 36px 22px 10px},clip]{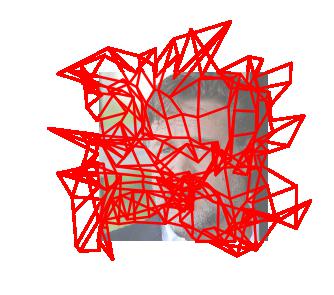} &
        \includegraphics[width=0.97\linewidth,trim={33px 24px 11px 10px},clip]{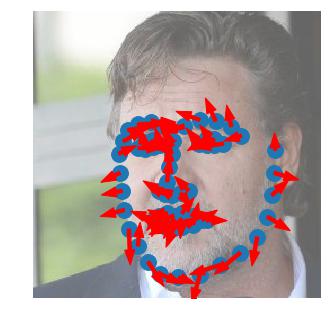} &
        \includegraphics[width=0.97\linewidth,trim={33px 24px 11px 10px},clip]{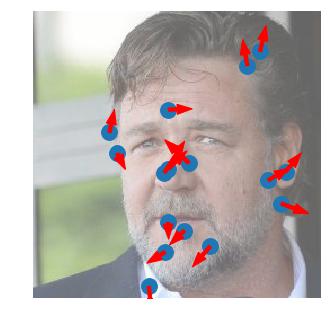} \\\midrule
        \raisebox{4em}{Image} &
        \includegraphics[width=0.97\linewidth,trim={33px 24px 11px 10px},clip]{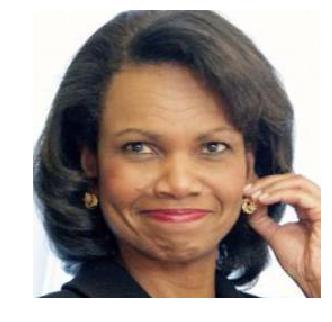} & 
        \includegraphics[width=0.97\linewidth,trim={33px 24px 11px 10px},clip]{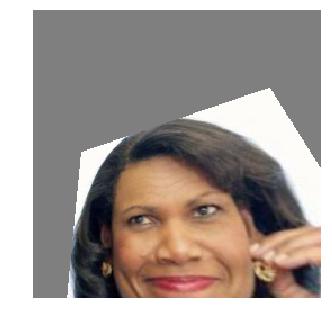} &
        \includegraphics[width=0.97\linewidth,trim={33px 24px 11px 10px},clip]{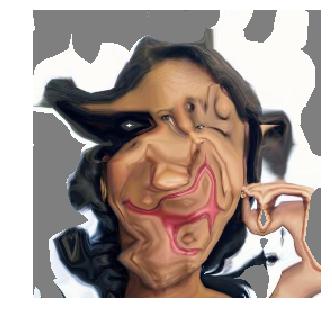} &
        \includegraphics[width=0.97\linewidth,trim={33px 24px 11px 10px},clip]{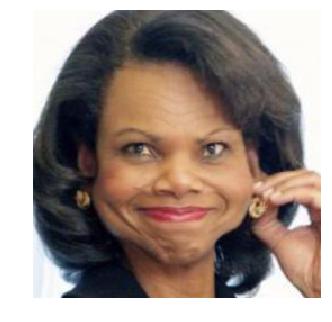} &
        \includegraphics[width=0.97\linewidth,trim={33px 24px 11px 10px},clip]{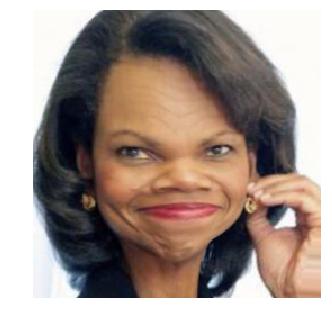} \\
        \raisebox{4em}{Transformation} &
        \includegraphics[width=0.97\linewidth,trim={33px 24px 11px 10px},clip]{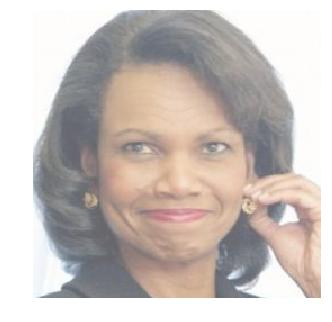} & 
        \includegraphics[width=0.97\linewidth,trim={33px 24px 11px 10px},clip]{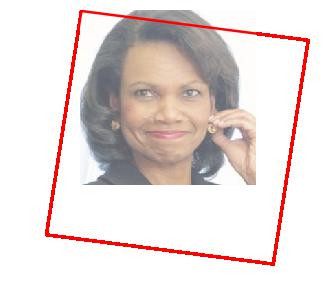} &
        \includegraphics[width=0.97\linewidth,trim={33px 36px 22px 10px},clip]{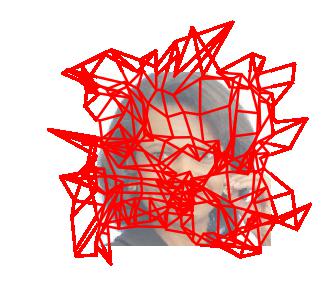} &
        \includegraphics[width=0.97\linewidth,trim={33px 24px 11px 10px},clip]{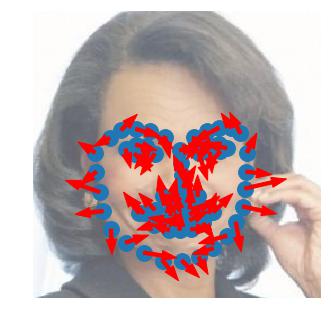} &
        \includegraphics[width=0.97\linewidth,trim={33px 24px 11px 10px},clip]{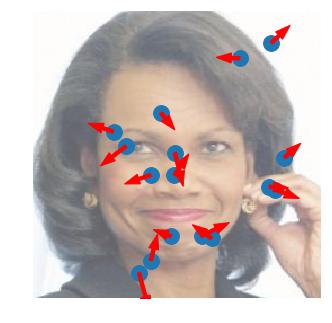} \\\midrule
    \end{tabularx}
    \caption{Different transformation methods. Input images are shown in the first column. The next four columns show the results and the transformation visualizations of four different models trained with different transformation methods. The landmark-based model uses $68$ landmarks detected by Dlib. Texture rendering is hidden here for clarity.}
    \label{appendix:fig:transformation}
\end{figure*}

To see the advantage of the proposed control-points estimation for automatic warping, we train three variants of our model by replacing the warping method with (1) projective transformation, (2) dense deformation and (3) landmark-based warping. In projective transformation, the warp controller outputs $8$ parameters for the transformation matrix. In dense deformation, the warp controller outputs a $16\times 16$ deformation grid, which is further interpolated into $256\times256$ for grid sampling. In landmark-based warping, we use the landmarks provided by Dlib\footnote{\url{http://dlib.net/face_landmark_detection.py.html}} and the warp controller only outputs the displacements. As shown in Figure~\ref{appendix:fig:transformation}, the warping is too limited in projective transformation for generating artistic caricatures and too unconstrained in dense deformation that it is difficult to train. Landmark-based warping yields reasonable results, but it is limited by the landmark detector. In comparison, our methods does not require any domain knowledge, has little limitation and leads to visually satisfying warping results.

\section{More Results}
\begin{figure*}[t]
\setlength\tabcolsep{0px}
\newcolumntype{Y}{>{\centering\arraybackslash}X}
    \centering
    \begin{tabularx}{\linewidth}{YYYYYYY}
        Input & w/o texture & w/o warping & w/o $\EL_g$  & w/o $\EL_p$ & w/o $\EL_{idt}$ & with all  \\
    \midrule
        \includegraphics[width=0.97\linewidth]{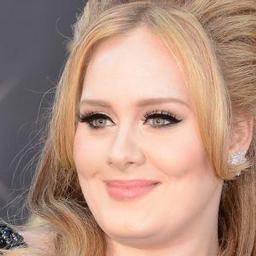} & 
        \includegraphics[width=0.97\linewidth]{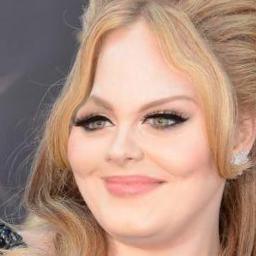} &
        \includegraphics[width=0.97\linewidth]{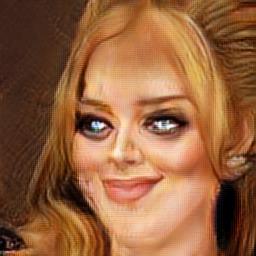} & 
        \includegraphics[width=0.97\linewidth]{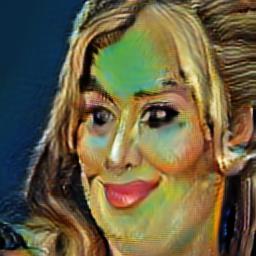} & 
        \includegraphics[width=0.97\linewidth]{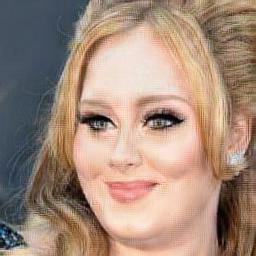} &
        \includegraphics[width=0.97\linewidth]{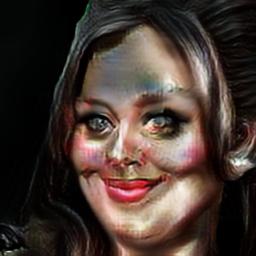} & 
        \includegraphics[width=0.97\linewidth]{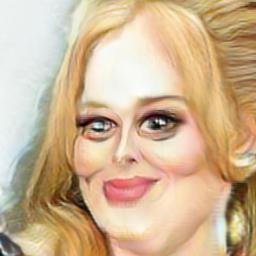} \\
        \includegraphics[width=0.97\linewidth]{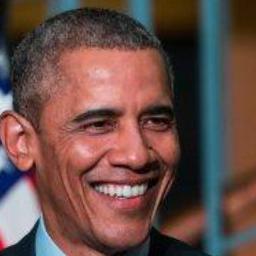} & 
        \includegraphics[width=0.97\linewidth]{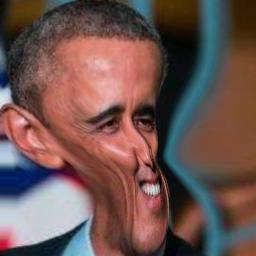} &
        \includegraphics[width=0.97\linewidth]{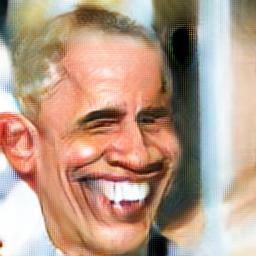} & 
        \includegraphics[width=0.97\linewidth]{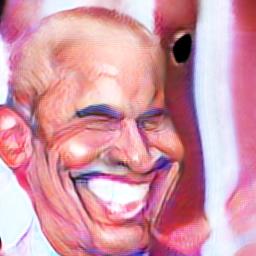} & 
        \includegraphics[width=0.97\linewidth]{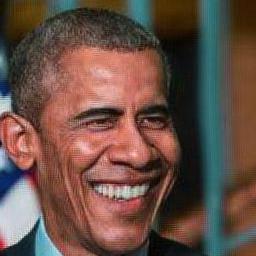} &
        \includegraphics[width=0.97\linewidth]{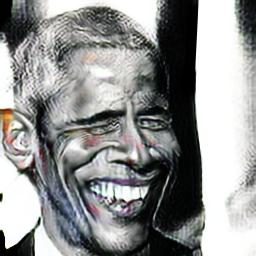} & 
        \includegraphics[width=0.97\linewidth]{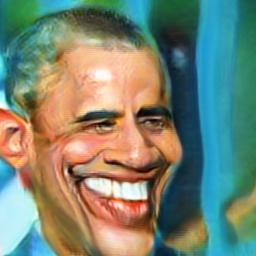} \\
        \includegraphics[width=0.97\linewidth]{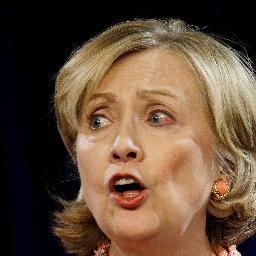} & 
        \includegraphics[width=0.97\linewidth]{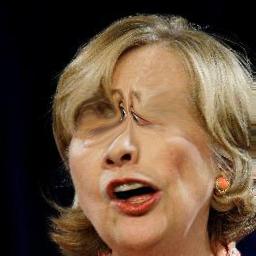} &
        \includegraphics[width=0.97\linewidth]{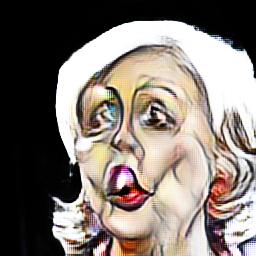} & 
        \includegraphics[width=0.97\linewidth]{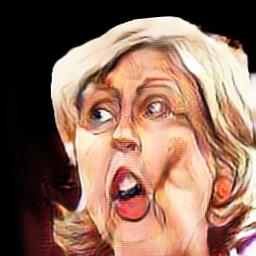} &
        \includegraphics[width=0.97\linewidth]{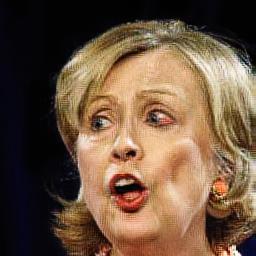} &
        \includegraphics[width=0.97\linewidth]{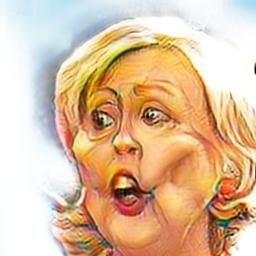} &
        \includegraphics[width=0.97\linewidth]{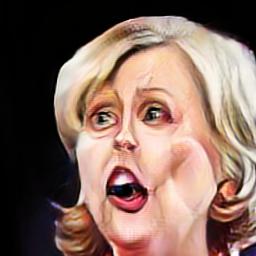} \\
        \includegraphics[width=0.97\linewidth]{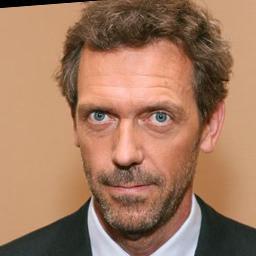} & 
        \includegraphics[width=0.97\linewidth]{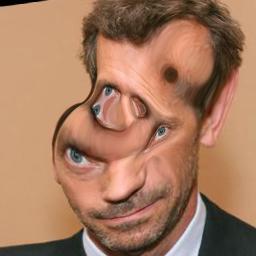} &
        \includegraphics[width=0.97\linewidth]{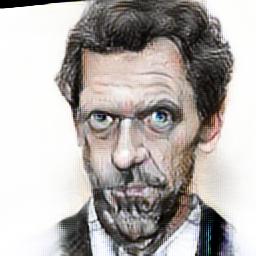} & 
        \includegraphics[width=0.97\linewidth]{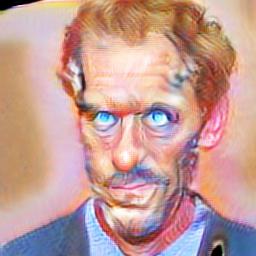} &
        \includegraphics[width=0.97\linewidth]{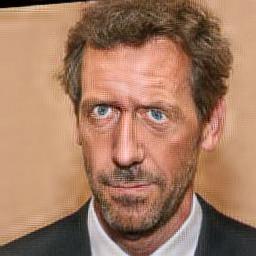} &
        \includegraphics[width=0.97\linewidth]{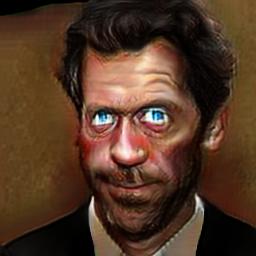} &
        \includegraphics[width=0.97\linewidth]{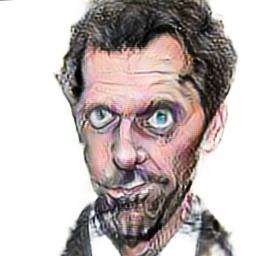} \\
        \includegraphics[width=0.97\linewidth]{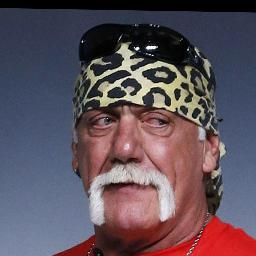} & 
        \includegraphics[width=0.97\linewidth]{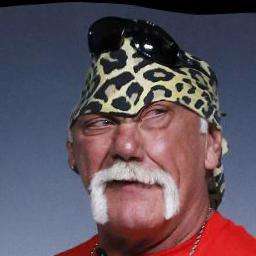} &
        \includegraphics[width=0.97\linewidth]{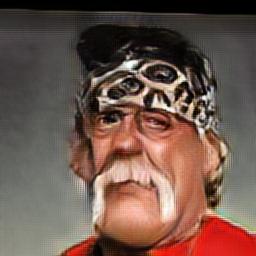} & 
        \includegraphics[width=0.97\linewidth]{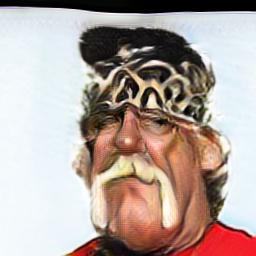} &
        \includegraphics[width=0.97\linewidth]{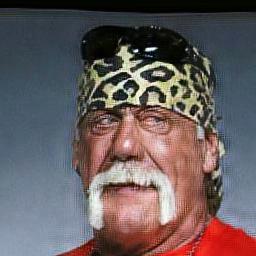} &
        \includegraphics[width=0.97\linewidth]{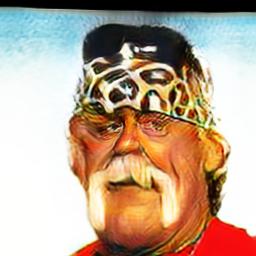} &
        \includegraphics[width=0.97\linewidth]{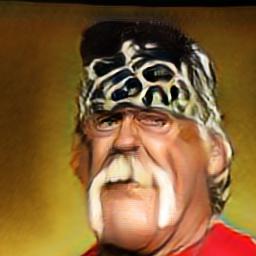} \\
        \includegraphics[width=0.97\linewidth]{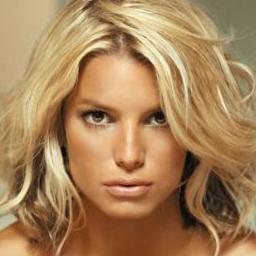} & 
        \includegraphics[width=0.97\linewidth]{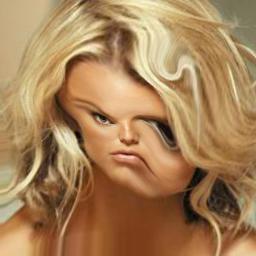} &
        \includegraphics[width=0.97\linewidth]{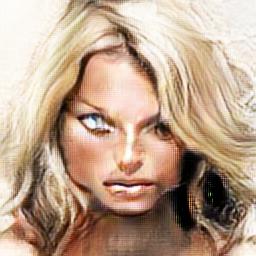} & 
        \includegraphics[width=0.97\linewidth]{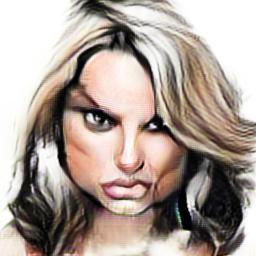} &
        \includegraphics[width=0.97\linewidth]{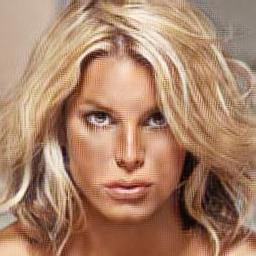} &
        \includegraphics[width=0.97\linewidth]{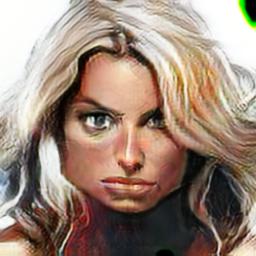} &
        \includegraphics[width=0.97\linewidth]{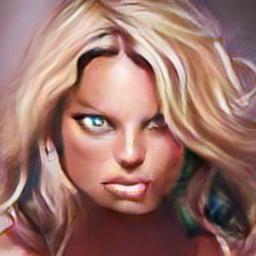} \\
        \includegraphics[width=0.97\linewidth]{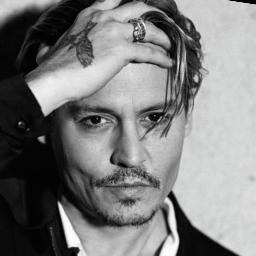} & 
        \includegraphics[width=0.97\linewidth]{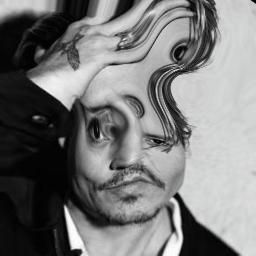} &
        \includegraphics[width=0.97\linewidth]{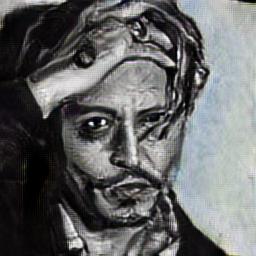} & 
        \includegraphics[width=0.97\linewidth]{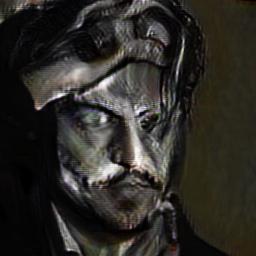} &
        \includegraphics[width=0.97\linewidth]{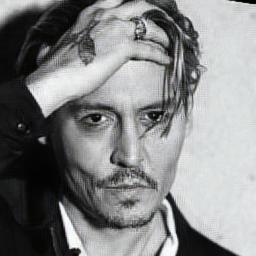} &
        \includegraphics[width=0.97\linewidth]{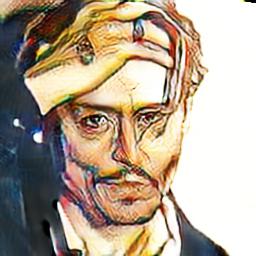} &
        \includegraphics[width=0.97\linewidth]{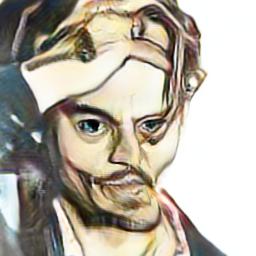} \\
        \includegraphics[width=0.97\linewidth]{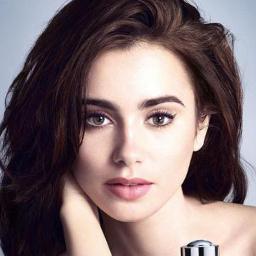} & 
        \includegraphics[width=0.97\linewidth]{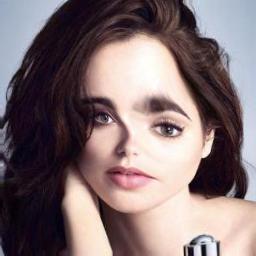} &
        \includegraphics[width=0.97\linewidth]{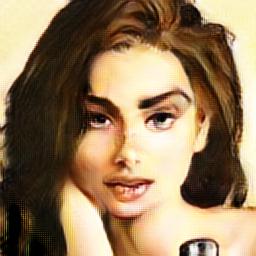} & 
        \includegraphics[width=0.97\linewidth]{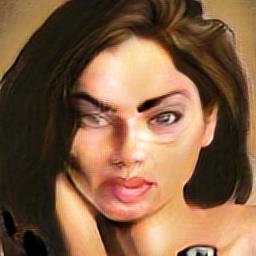} &
        \includegraphics[width=0.97\linewidth]{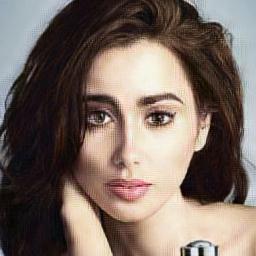} &
        \includegraphics[width=0.97\linewidth]{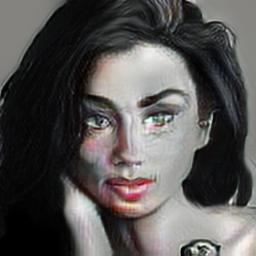} &
        \includegraphics[width=0.97\linewidth]{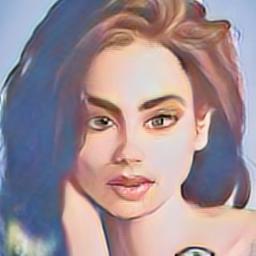} \\
    \bottomrule
    \end{tabularx}
    \caption{More results on ablation study. Input images are shown in the first column. The subsequent columns show the results of different models trained without a certain module or loss. The texture style codes are randomly sampled from the normal distribution.}
    \label{appendix:fig:ablation}
\end{figure*}

\paragraph{Ablation Study} We show more results of the ablation study in Figure~\ref{appendix:fig:ablation}. The results are consistent with those in the main paper: (1) the joint learning of texture rendering and warping are crucial for generating realistic caricature images and (2) without patch adversarial loss or identity-preservation adversarial loss, the model cannot learn to generate caricatures with various texture styles and shape exaggeration styles.

\begin{figure*}[t]
\setlength\tabcolsep{0px}
\newcolumntype{Y}{>{\centering\arraybackslash}X}
    \centering
    \begin{tabularx}{\linewidth}{YYYYYY}
        Input & code $1$ & code $2$  & code $3$ &code $4$ & code $5$  \\
    \midrule
        \includegraphics[width=0.97\linewidth]{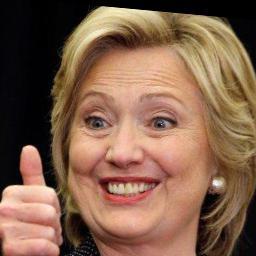} &
        \includegraphics[width=0.97\linewidth]{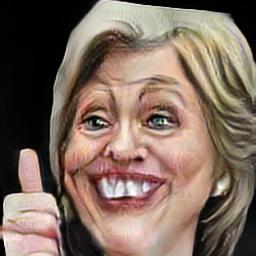} & 
        \includegraphics[width=0.97\linewidth]{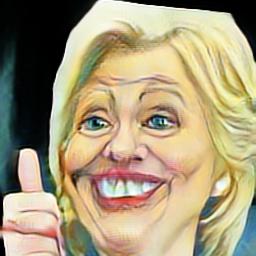} &
        \includegraphics[width=0.97\linewidth]{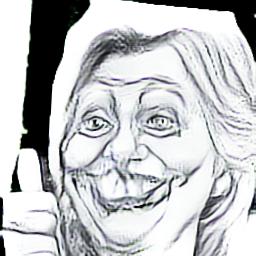} &
        \includegraphics[width=0.97\linewidth]{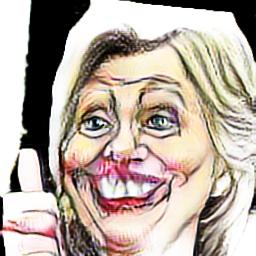} &
        \includegraphics[width=0.97\linewidth]{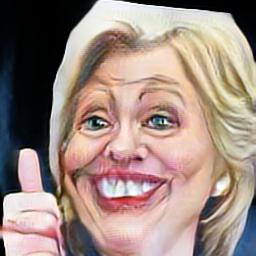} \\
        \includegraphics[width=0.97\linewidth]{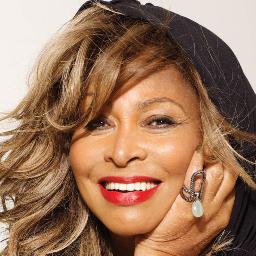} &
        \includegraphics[width=0.97\linewidth]{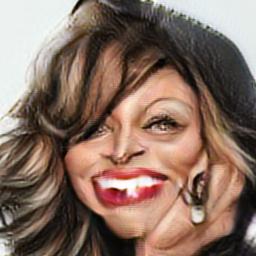} & 
        \includegraphics[width=0.97\linewidth]{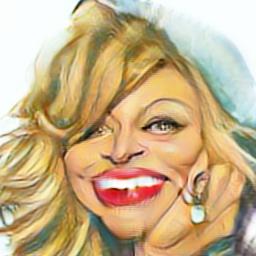} &
        \includegraphics[width=0.97\linewidth]{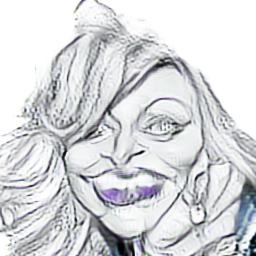} &
        \includegraphics[width=0.97\linewidth]{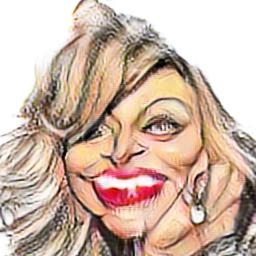} &
        \includegraphics[width=0.97\linewidth]{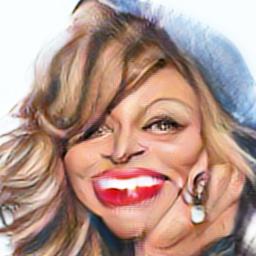} \\
        \includegraphics[width=0.97\linewidth]{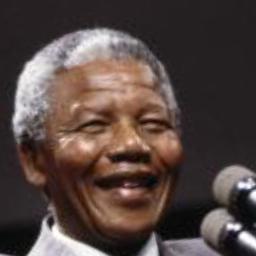} &
        \includegraphics[width=0.97\linewidth]{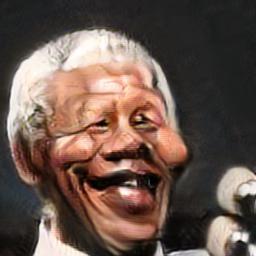} & 
        \includegraphics[width=0.97\linewidth]{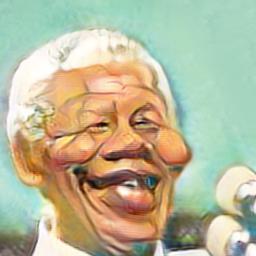} &
        \includegraphics[width=0.97\linewidth]{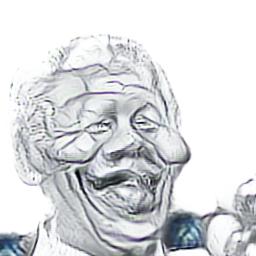} &
        \includegraphics[width=0.97\linewidth]{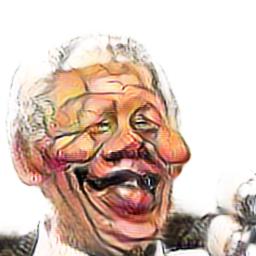} &
        \includegraphics[width=0.97\linewidth]{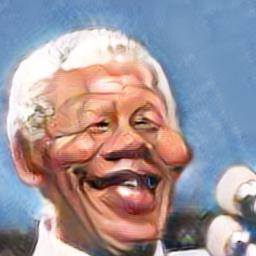} \\
        \includegraphics[width=0.97\linewidth]{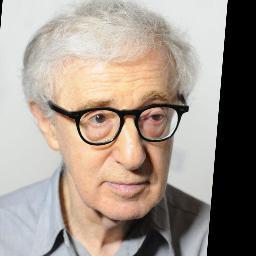} &
        \includegraphics[width=0.97\linewidth]{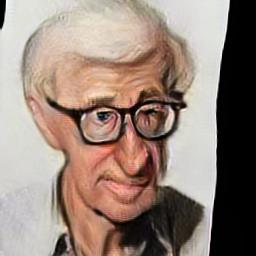} & 
        \includegraphics[width=0.97\linewidth]{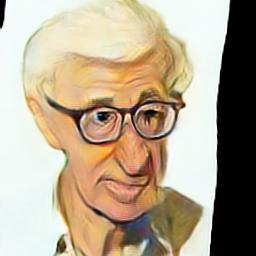} &
        \includegraphics[width=0.97\linewidth]{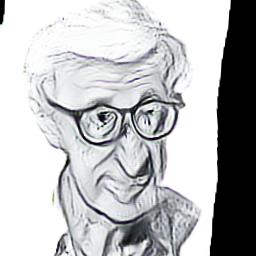} &
        \includegraphics[width=0.97\linewidth]{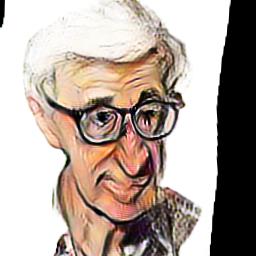} &
        \includegraphics[width=0.97\linewidth]{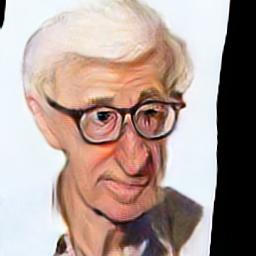} \\
        \includegraphics[width=0.97\linewidth]{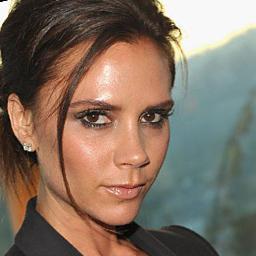} &
        \includegraphics[width=0.97\linewidth]{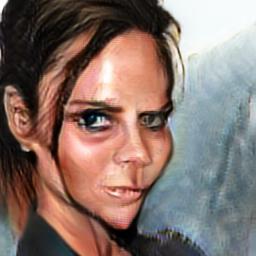} & 
        \includegraphics[width=0.97\linewidth]{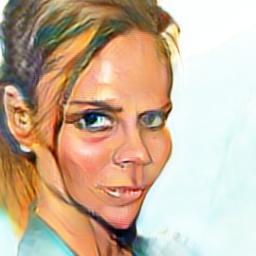} &
        \includegraphics[width=0.97\linewidth]{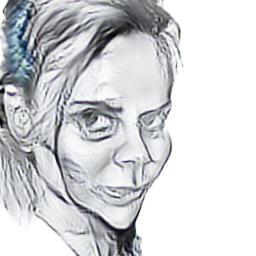} &
        \includegraphics[width=0.97\linewidth]{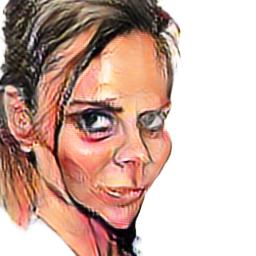} &
        \includegraphics[width=0.97\linewidth]{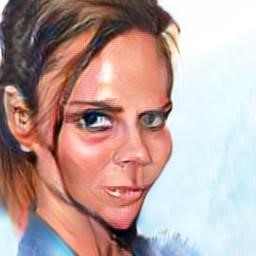} \\
        \includegraphics[width=0.97\linewidth]{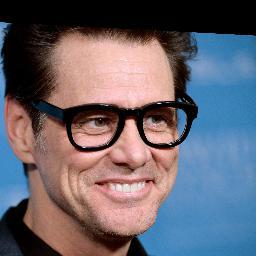} &
        \includegraphics[width=0.97\linewidth]{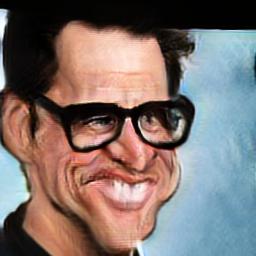} & 
        \includegraphics[width=0.97\linewidth]{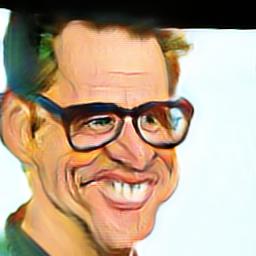} &
        \includegraphics[width=0.97\linewidth]{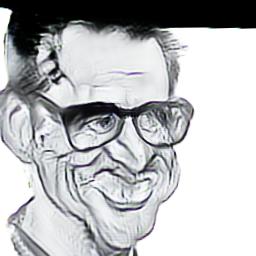} &
        \includegraphics[width=0.97\linewidth]{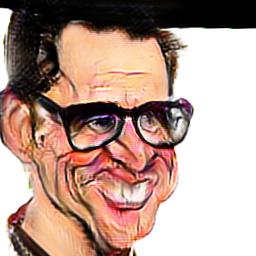} &
        \includegraphics[width=0.97\linewidth]{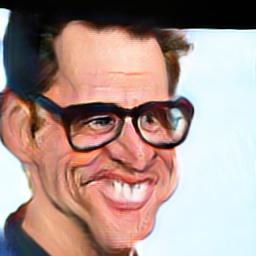} \\
        \includegraphics[width=0.97\linewidth]{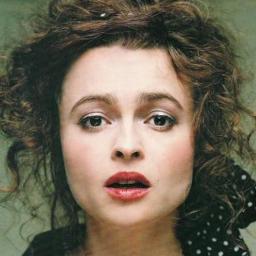} &
        \includegraphics[width=0.97\linewidth]{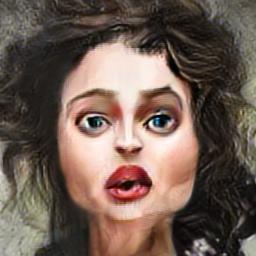} & 
        \includegraphics[width=0.97\linewidth]{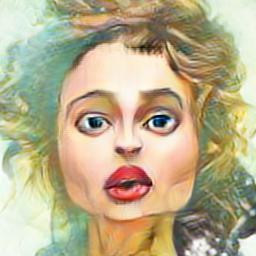} &
        \includegraphics[width=0.97\linewidth]{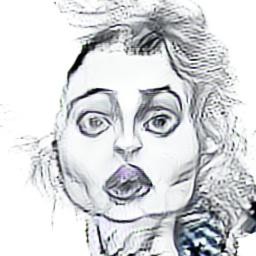} &
        \includegraphics[width=0.97\linewidth]{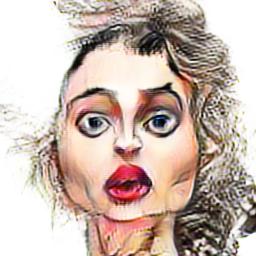} &
        \includegraphics[width=0.97\linewidth]{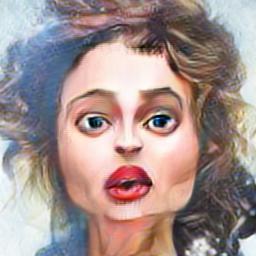} \\
    \bottomrule
    \end{tabularx}
    \caption{Results of five different texture styles. Input images are shown in the first column. Subsequent five columns show the results of WarpGAN using five style codes sampled randomly from the normal distribution. All the images in the same column are generated with the same latent style code.}
    \label{appendix:fig:styles}
\end{figure*}

\paragraph{Different Texture Styles} More results of texture style controlling are shown in Figure~\ref{appendix:fig:styles}. Five latent style codes are randomly sampled from the normal distribution $\mathcal{N}(0,\mathbf{I})$. Images in the same column in Figure~\ref{appendix:fig:styles} are generated with the same style code. 

\begin{figure*}[t]
\setlength\tabcolsep{0px}
\newcolumntype{Y}{>{\centering\arraybackslash}X}
    \centering
    \begin{tabularx}{\linewidth}{YYYY}
        \includegraphics[width=0.49\linewidth]{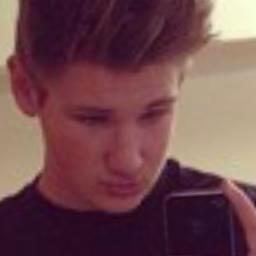}\includegraphics[width=0.49\linewidth]{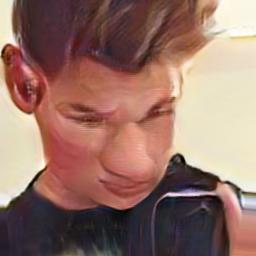} &
        \includegraphics[width=0.49\linewidth]{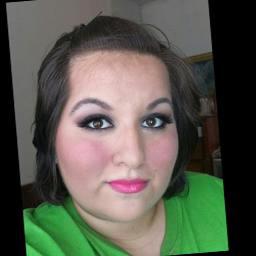}\includegraphics[width=0.49\linewidth]{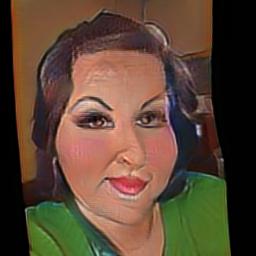} &
        \includegraphics[width=0.49\linewidth]{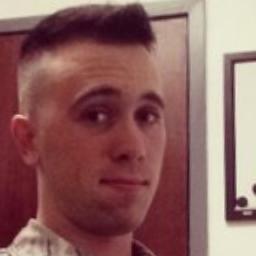}\includegraphics[width=0.49\linewidth]{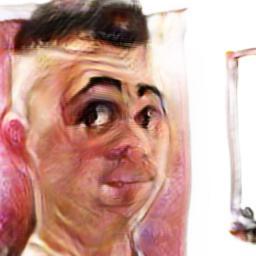} &
        \includegraphics[width=0.49\linewidth]{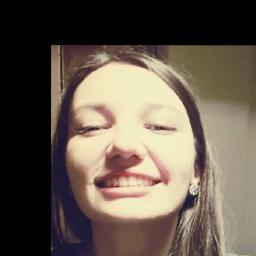}\includegraphics[width=0.49\linewidth]{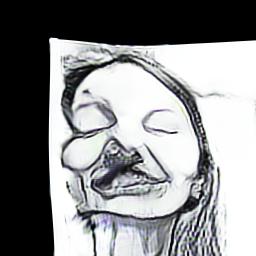} \\
        \includegraphics[width=0.49\linewidth]{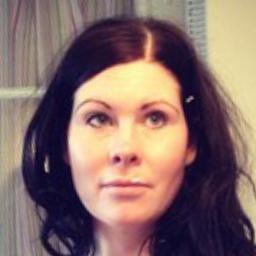}\includegraphics[width=0.49\linewidth]{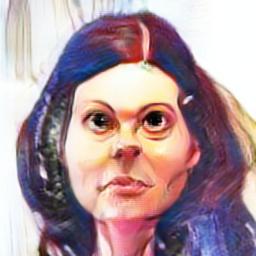} &
        \includegraphics[width=0.49\linewidth]{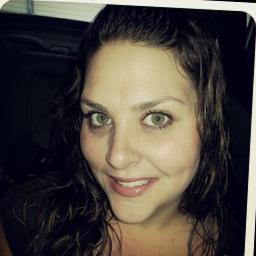}\includegraphics[width=0.49\linewidth]{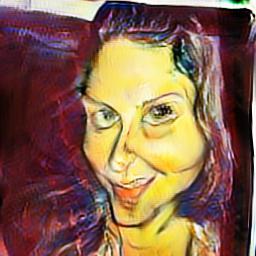} &
        \includegraphics[width=0.49\linewidth]{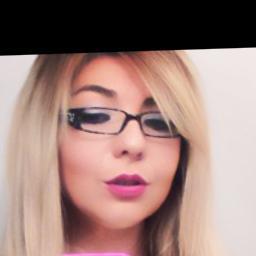}\includegraphics[width=0.49\linewidth]{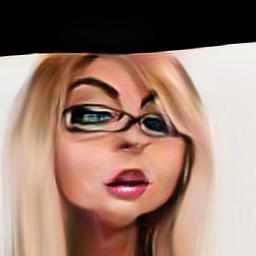} &
        \includegraphics[width=0.49\linewidth]{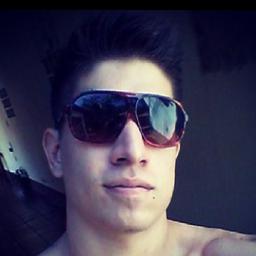}\includegraphics[width=0.49\linewidth]{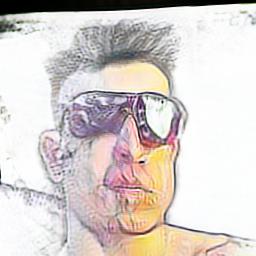} \\
        \includegraphics[width=0.49\linewidth]{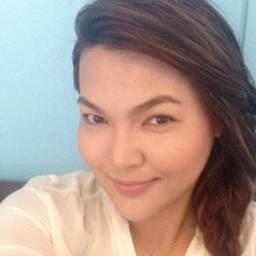}\includegraphics[width=0.49\linewidth]{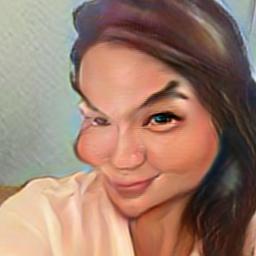} &
        \includegraphics[width=0.49\linewidth]{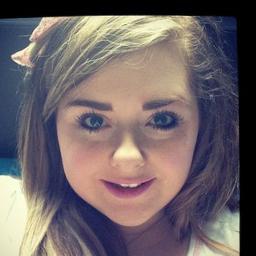}\includegraphics[width=0.49\linewidth]{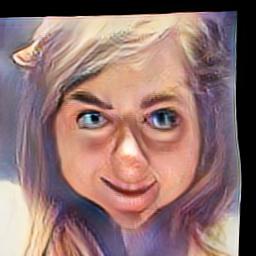} &
        \includegraphics[width=0.49\linewidth]{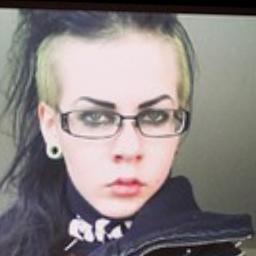}\includegraphics[width=0.49\linewidth]{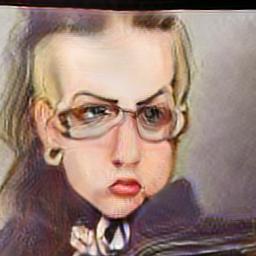} &
        \includegraphics[width=0.49\linewidth]{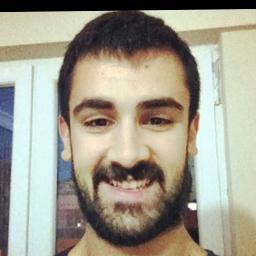}\includegraphics[width=0.49\linewidth]{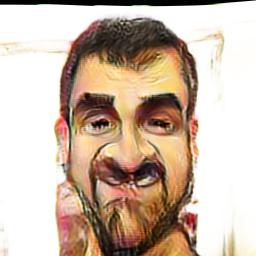}\\ 
        \includegraphics[width=0.49\linewidth]{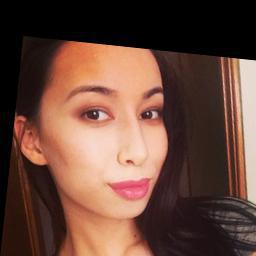}\includegraphics[width=0.49\linewidth]{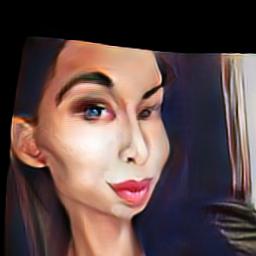} &
        \includegraphics[width=0.49\linewidth]{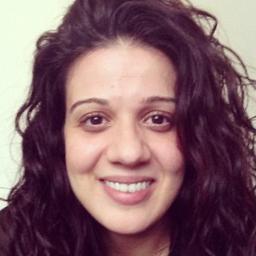}\includegraphics[width=0.49\linewidth]{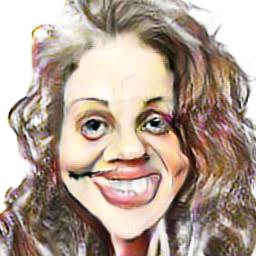} &
        \includegraphics[width=0.49\linewidth]{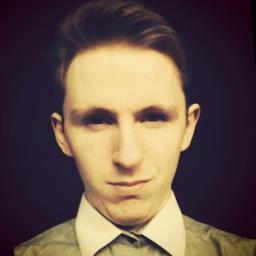}\includegraphics[width=0.49\linewidth]{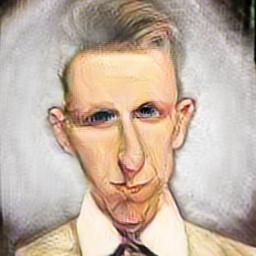} &
        \includegraphics[width=0.49\linewidth]{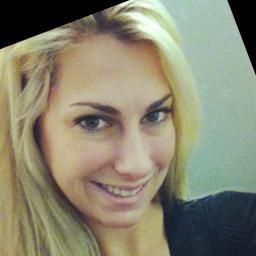}\includegraphics[width=0.49\linewidth]{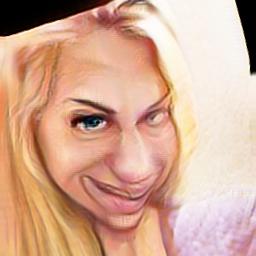} \\
        \includegraphics[width=0.49\linewidth]{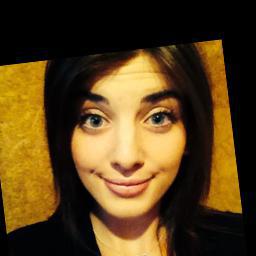}\includegraphics[width=0.49\linewidth]{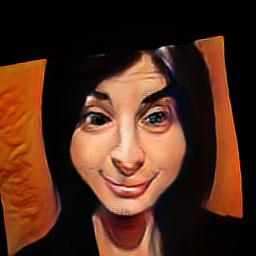} &
        \includegraphics[width=0.49\linewidth]{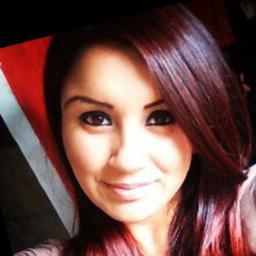}\includegraphics[width=0.49\linewidth]{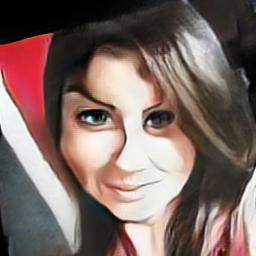} &
        \includegraphics[width=0.49\linewidth]{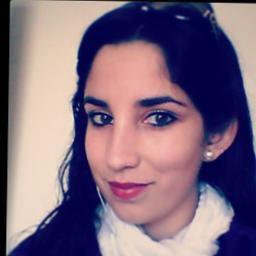}\includegraphics[width=0.49\linewidth]{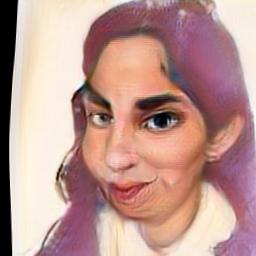} &
        \includegraphics[width=0.49\linewidth]{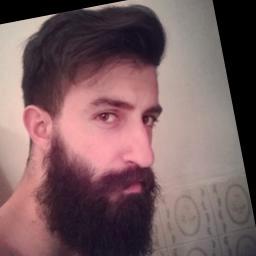}\includegraphics[width=0.49\linewidth]{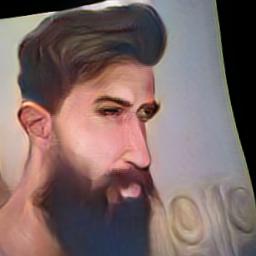} \\
        \includegraphics[width=0.49\linewidth]{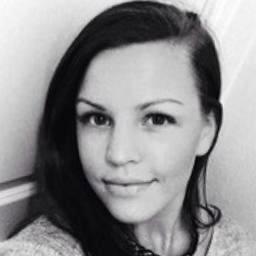}\includegraphics[width=0.49\linewidth]{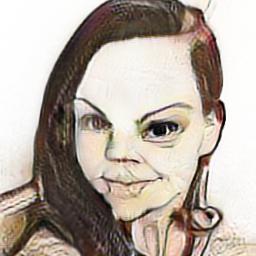} &
        \includegraphics[width=0.49\linewidth]{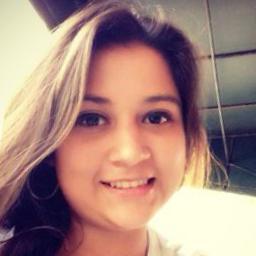}\includegraphics[width=0.49\linewidth]{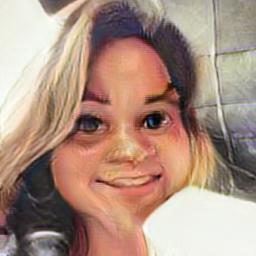} &
        \includegraphics[width=0.49\linewidth]{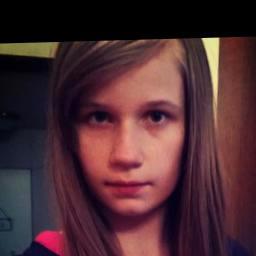}\includegraphics[width=0.49\linewidth]{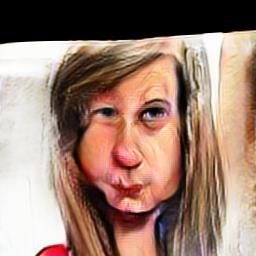} &
        \includegraphics[width=0.49\linewidth]{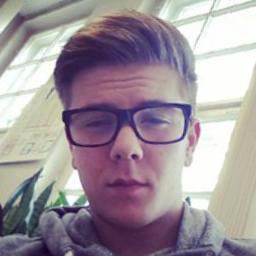}\includegraphics[width=0.49\linewidth]{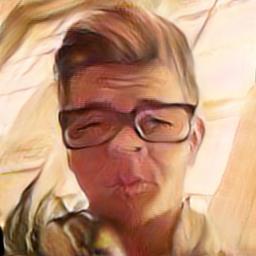} \\
        \includegraphics[width=0.49\linewidth]{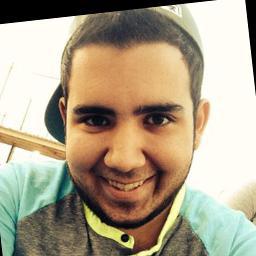}\includegraphics[width=0.49\linewidth]{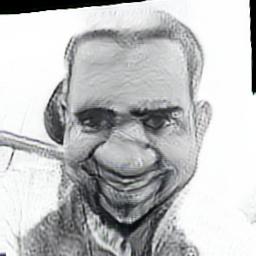} &
        \includegraphics[width=0.49\linewidth]{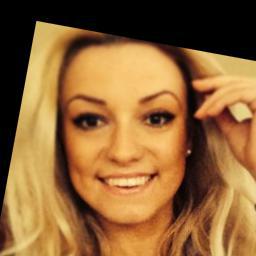}\includegraphics[width=0.49\linewidth]{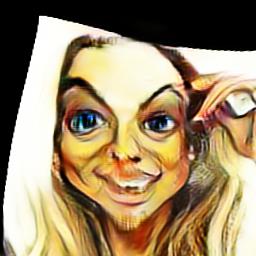} &
        \includegraphics[width=0.49\linewidth]{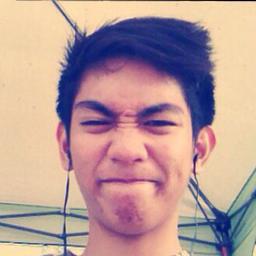}\includegraphics[width=0.49\linewidth]{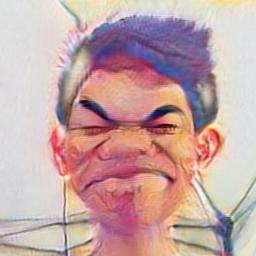} &
        \includegraphics[width=0.49\linewidth]{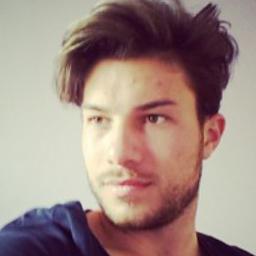}\includegraphics[width=0.49\linewidth]{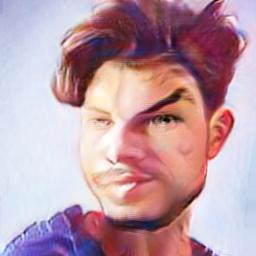} \\
        \includegraphics[width=0.49\linewidth]{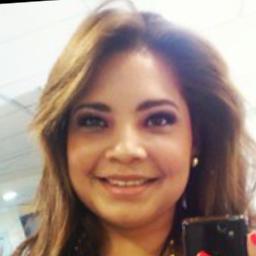}\includegraphics[width=0.49\linewidth]{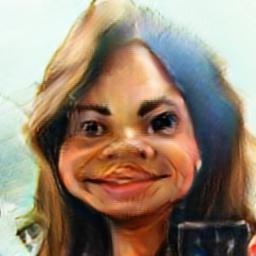} &
        \includegraphics[width=0.49\linewidth]{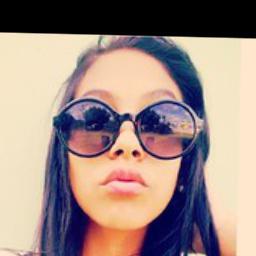}\includegraphics[width=0.49\linewidth]{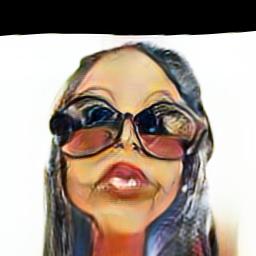} &
        \includegraphics[width=0.49\linewidth]{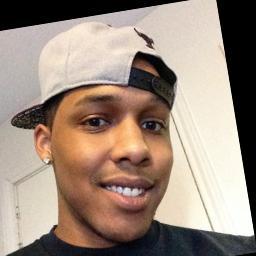}\includegraphics[width=0.49\linewidth]{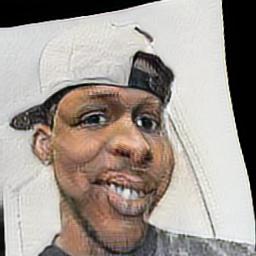} &
        \includegraphics[width=0.49\linewidth]{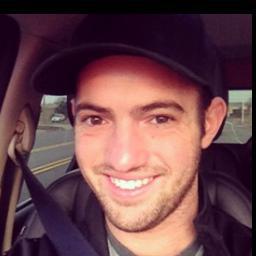}\includegraphics[width=0.49\linewidth]{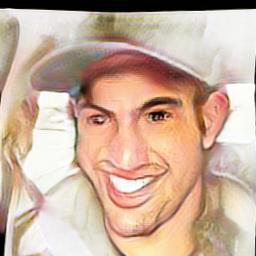} \\
        \includegraphics[width=0.49\linewidth]{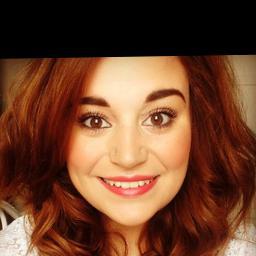}\includegraphics[width=0.49\linewidth]{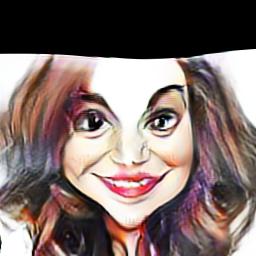} &
        \includegraphics[width=0.49\linewidth]{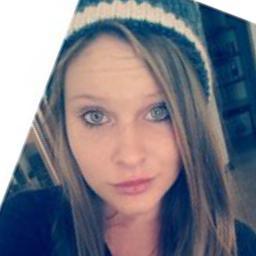}\includegraphics[width=0.49\linewidth]{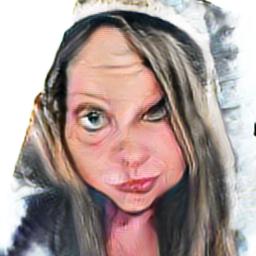} &
        \includegraphics[width=0.49\linewidth]{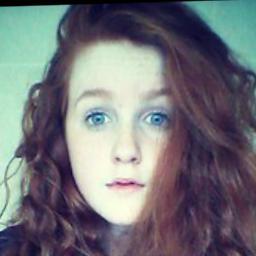}\includegraphics[width=0.49\linewidth]{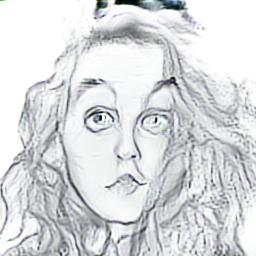} &
        \includegraphics[width=0.49\linewidth]{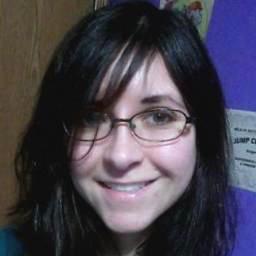}\includegraphics[width=0.49\linewidth]{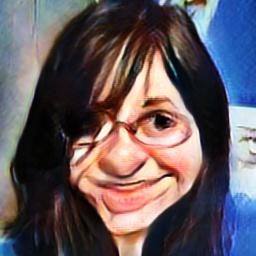} \\
    \end{tabularx}
    % \vspace{-1.0em}
    \caption{Example results on the Selfie dataset. This is a cross-dataset evaluation and no training is involved. In each pair, the left image is the input and the right image is the output of WarpGAN with a random texture style.}
    \label{appendix:fig:selfie}
\end{figure*}

\paragraph{Selfie Dataset} To test the performance of our model in more application scenarios, we download the public Selfie dataset\footnote{\url{http://crcv.ucf.edu/data/Selfie/}}~\cite{kalayeh2015take} for cross-dataset evaluation. The dataset includes $46,836$ public selfies crawled from Internet. Unlike our training dataset (WebCaricature), the identities in this dataset are not restricted to celebrities and there is a difference between the visual styles of these images and the ones in our training dataset. The results are shown in Figure~\ref{appendix:fig:selfie}. 

\end{document}